\def\paperlanguage{en} 

\documentclass[letterpaper, 10 pt, conference]{ieeeconf}  

\usepackage{bm}
\usepackage{cite}
\usepackage{flushend}
\include{preamble}
\usepackage{booktabs}

\IEEEoverridecommandlockouts                              

\title{\LARGE \textbf
  {
    \switchlanguage%
    {%
			KLEIYN : A Quadruped Robot with an Active Waist \\ for Both Locomotion and Wall Climbing
		}%
    {%
			KLEIYN : 腰関節を有し歩行と壁面移動を両立する四脚ロボット
    }%
  }
}

\author{Keita Yoneda$^{1}$, Kento Kawaharazuka$^{1, 2}$, Temma Suzuki$^{1}$, Takahiro Hattori$^{1}$, Kei Okada$^{1}$
  \thanks{$^{1}$ The authors are with the Department of Mechano-Informatics, Graduate School of Information Science and Technology, The University of Tokyo, 7-3-1 Hongo, Bunkyo-ku, Tokyo, 113-8656, Japan.
    {\texttt\small [yoneda, kawaharazuka, t-suzuki, t-hattori, k-okada]@jsk.imi.i.u-tokyo.ac.jp}
  }
	\thanks{$^{2}$ The author is with the AI Center, Graduate School of Information Science and Technology, The University of Tokyo, Japan.
	}
}
\begin{document}

\maketitle
\thispagestyle{empty}
\pagestyle{empty}

\begin{abstract}
  \switchtext%
  {%
	In recent years, advancements in hardware have enabled quadruped robots to operate with high power and speed, while robust locomotion control using reinforcement learning (RL) has also been realized.
	As a result, expectations are rising for the automation of tasks such as material transport and exploration in unknown environments.
	However, autonomous locomotion in rough terrains with significant height variations requires vertical movement, and robots capable of performing such movements stably, along with their control methods, have not yet been fully established.
	In this study, we developed the quadruped robot KLEIYN, which features a waist joint, and aimed to expand quadruped locomotion by enabling chimney climbing through RL.
	To facilitate the learning of vertical motion, we introduced Contact-Guided Curriculum Learning (CGCL).
	As a result, KLEIYN successfully climbed walls ranging from 800 mm to 1000 mm in width at an average speed of 150 mm/s, 50 times faster than conventional robots.
	Furthermore, we demonstrated that the introduction of a waist joint improves climbing performance, particularly enhancing tracking ability on narrow walls.
	}%
  {%
	近年, 大出力かつ高速に動作可能な四脚ロボットのハードウェアと, 強化学習を利用したロバストな歩行制御が実現されており, 未知環境における物資運搬や探索活動の自動化などが期待されている. 
	岩場など高低差の激しい環境における自律移動では, 垂直方向の移動も必要となる.  しかしこれを安定して行えるロボットやその制御手法は明らかになっていない.
	本研究では, 腰関節を有する四脚ロボット KLEIYN を開発し, 強化学習により壁突っ張りを利用した壁面移動を実現することで, 四脚ロボットの移動形態の拡張を行う.
	垂直方向動作の学習を促進するように地形を変化させる, 接触誘導型カリキュラム学習を行った.
	これにより, KLEIYN は幅800 mmから1000 mmの多様な壁を従来のロボットの約50 倍である平均速度 150 mm/s で登ることに成功した.
	また腰関節を導入することで壁登り動作の性能が向上することを示し, 特に狭い幅の壁において追従性が大きく向上することを確認した.
	}%
	{%
	}%
\end{abstract}

\section{Introduction}\label{sec:introduction}
\switchlanguage%
{%
	In recent years, the development of quadruped robots has been actively conducted \cite{hutter2016anymal, katz2019minicheetah, unitree2022go1}.
	These robots utilize high-power, low-gear-ratio motors, which enable dynamic motions such as jumping \cite{bellegarda2024robust}.
	Furthermore, RL in simulators \cite{hwangbo2019anymal} has facilitated the realization of unified locomotion control that adapts to various ground conditions, including steps and slippery surfaces \cite{tsounis2020deepgait, aractingi2023controlling, duan2024cassie}.
	The evolution of both hardware and software has significantly expanded the capabilities of quadruped robots, leading to expectations for various applications. One such application is the automation of material transport and exploration activities in disaster sites and unexplored natural environments \cite{miki2022anymal}. These environments are characterized by large obstacles, such as collapsed buildings and rocks, with significant height variations.
	However, while quadruped robots have demonstrated effective horizontal locomotion, robots capable of both stable vertical movement and horizontal locomotion have yet to be realized. Although some quadruped robots specialized for vertical movement have been developed, their body structures are highly specialized for climbing, making horizontal movement difficult \cite{parness2017lemur3}.
	To address this limitation, we aim to establish a quadruped robot architecture that integrates both horizontal and vertical locomotion.
	Specifically, we develop KLEIYN, a quadruped robot capable of both walking and wall climbing, as shown in \figref{fig:kleiyn}.

	The key contributions of this study are summarized as follows:

	\begin{enumerate}
		\item Development of a quadruped robot, KLEIYN, featuring a waist joint and quasi-direct-drive joints.
		\item Proposal of Contact-Guided Curriculum Learning (CGCL), which gradually modifies the wall-floor transition from a curved surface to a vertical one.
		\item Realization of chimney climbing on KLEIYN.
		\item Demonstration of the effectiveness of the waist joint in chimney climbing motion.
	\end{enumerate}

	\begin{figure}[t]
    \centering
      \includegraphics[width=0.95\columnwidth]{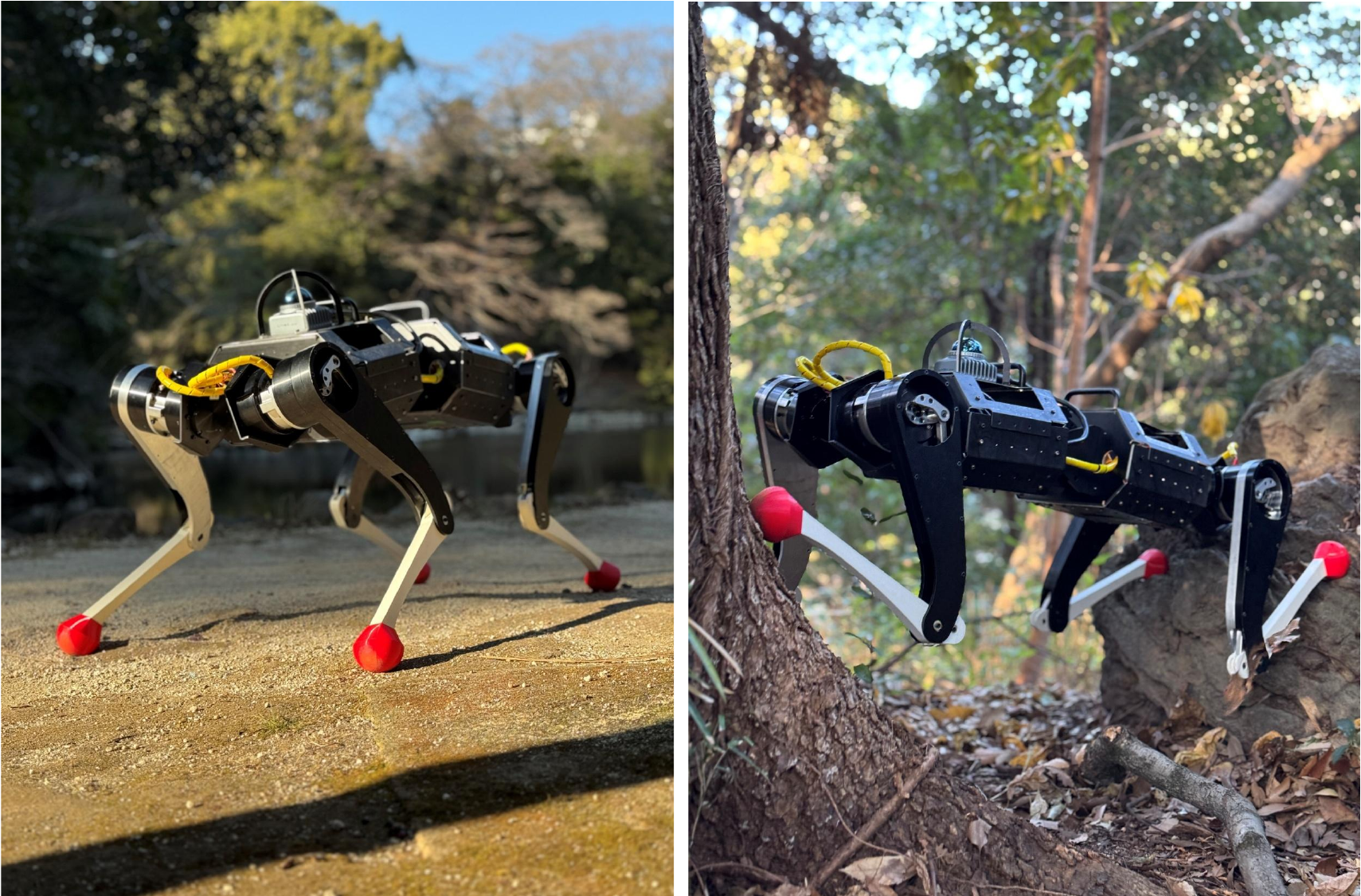}
      \vspace{-1.0ex}
      \caption{Overview of KLEIYN, the quadruped robot with an active waist joint developed in this study. It is capable of both walking and chimney climbing for wide range of walls.}
      \vspace{-3.0ex}
      \label{fig:kleiyn}
    
  \end{figure}
}%
{%
	近年四脚ロボットの開発及び制御法の研究が盛んに行われている\cite{hutter2016anymal, katz2019minicheetah, unitree2022go1, grimminger2020open}.
	これらはアクチュエータとして大出力かつ高速に動作可能な低減速比モータを用いることでバックドライバビリティを向上させ,跳躍などのダイナミックな動作を可能にしている\cite{bellegarda2024robust}.
	またシミュレータを用いた強化学習\cite{hwangbo2019anymal}により段差や滑りやすい地面などの多様な地面状況に対する統一的な歩行制御が実現された\cite{tsounis2020deepgait,aractingi2023controlling,duan2024cassie}. 
	このハードウェアとソフトウェア両面の進化により四脚ロボットで実現可能な動作は大きく広がり,様々な応用が期待されている.
	応用先の一つとして災害現場や未探索の自然環境での物資運搬,探索活動の自動化が進められている\cite{miki2022anymal}.
	これらの環境は倒壊した建物や岩などの大きな障害物が多く存在し高低差が激しいのが特徴である.
	しかし水平方向に移動可能な四脚ロボットで垂直移動可能なロボットは未だ実現されていない.
	また垂直方向のみ移動可能な四脚ロボットは開発されているものの,身体構造がその動作に特化しており,水平方向の移動が困難である\cite{parness2017lemur3}.
	そこで本研究では \figref{fig:kleiyn}に示す歩行と壁面移動を両立する四脚ロボットKLEIYNの開発を通して水平移動と垂直移動を両立する四脚ロボットの構成法を明らかにすることを目指した.

	本研究で新しく行ったことは以下のようにまとめられる.
	\begin{enumerate}
    \item 腰関節を有する四脚ロボットKLEIYNを開発した
    \item 壁登りを効率的に学習するために壁と床の接合部をなめらかな曲面から垂直な形状に徐々に変化させる学習手法を提案した. 
    \item 突っ張りを利用した壁登り動作を実機で実現し,間隔が800 mm から 1000 mmの多様な壁に対し壁登りが可能であることを示した. 
    \item 腰関節がなければ登れない狭い幅の壁を登ることで腰関節の有用性を示した.
    \item 腰関節を導入することで狭い幅の壁を登る際に脚関節で必要なトルクが減少することを確認した.
	\end{enumerate}
}%
\section{Related Works}\label{sec:related_works}
\subsection{Wall Climbing on Legged Robots}\label{subsec:climbing-robots}
\switchtext%
{%
	Wall climbing requires the robot to move against gravity, and the robot must support its body using the wall to prevent falling.

	Focusing on how the body is supported, wall-climbing legged robots can be classified into two categories: face climbing robots, which climb by gripping the wall, and chimney climbing robots, which climb by pressing against the wall.  
	Examples of face climbing robots include LEMUR 3\cite{parness2017lemur3}, SCALER\cite{tanaka2022scaler}, and LORIS\cite{nadan2024loris}.  
	These robots use grippers to grasp wall protrusions, allowing them to support their bodies on vertical surfaces.  
	However, these robots are unable to climb smooth walls that lack graspable protrusions.  
	Additionally, since grasping-based body support is not required for walking on flat ground, the grippers on the feet impede the walking.

	On the other hand, both chimney climbing and normal locomotion involve pushing against the wall or floor with the feet, making them achievable with the same hardware.
	Moreover, chimney climbing locomotion can be performed if two opposing walls are present.
	Therefore, for a quadruped robot aiming to integrate both walking and wall climbing, chimney climbing is a suitable approach.
	Consequently, this study aims to develop a robot capable of chimney climbing, and the term "wall climbing" in this context specifically refers to chimney climbing.

	However, no quadruped robot has been developed that achieves vertical movement by bracing against walls.  
	The only legged robot capable of this is the hexapod robot SiLVIA\cite{lin2018twowall, lin2019twowall}.  

	SiLVIA is capable of chimney climbing at a speed of 3 mm/s.
	However, SiLVIA has several limitations.
	First, its slow leg movements result in a low climbing speed.
	Second, since its control primarily relies on following a predefined leg trajectory, it lacks the ability to recover from foot slippage.
	Finally, the absence of joints in the torso prevents it from climbing walls narrower than the width of its body.

	\begin{figure}[t]
		\centering
		\includegraphics[width=0.95\columnwidth]{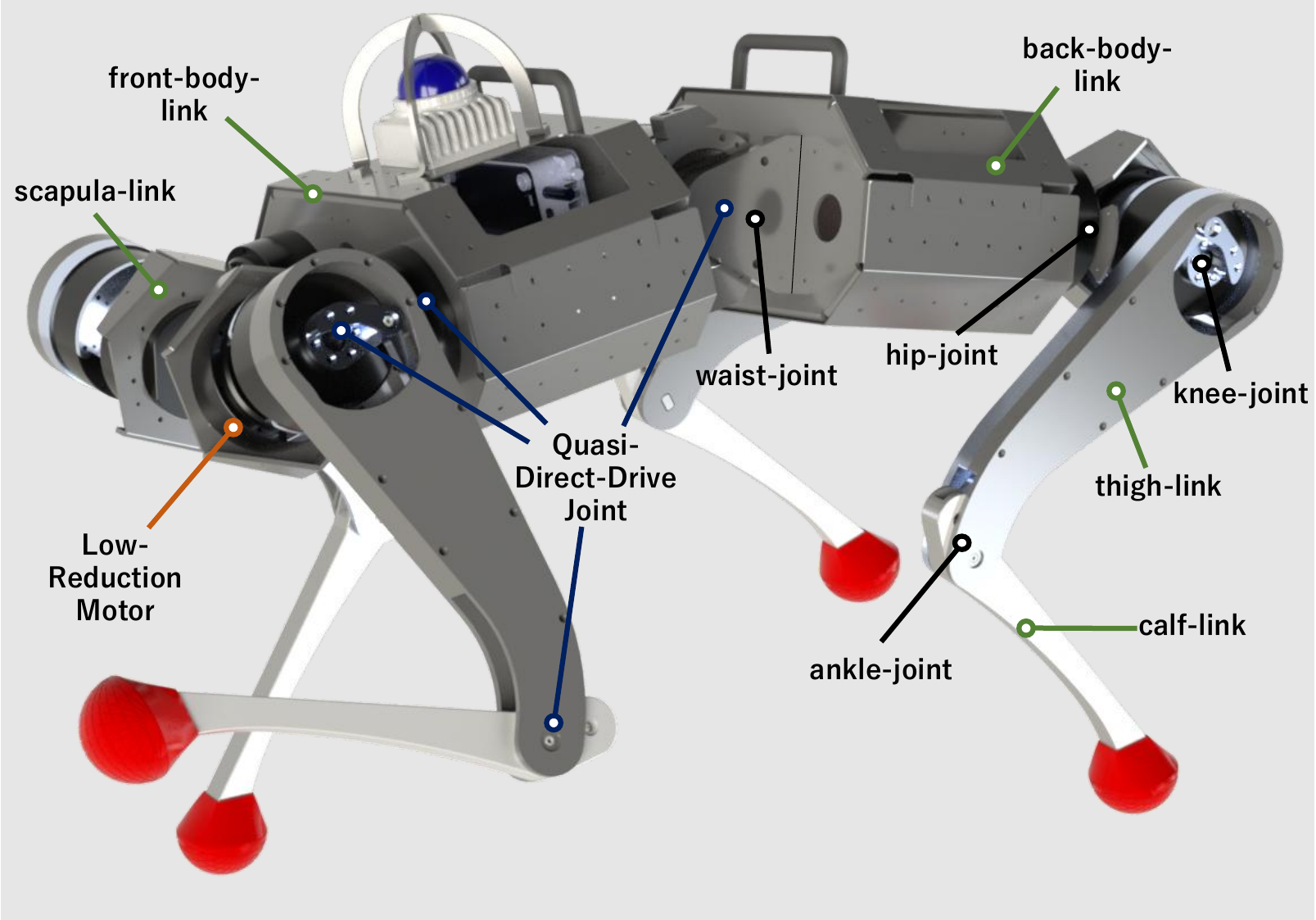}
		\vspace{-1.0ex}
		\caption{The design of KLEIYN: Each leg has 3 degrees of freedom (DOF), and a pitch-axis joint with 1-DOF is incorporated into the waist. Each joint is driven by a low-reduction-ratio motor using a quasi-direct drive mechanism.}
		\vspace{-3.0ex}
		\label{fig:kleiyn-design-overview}
	\end{figure}
}%
{%
	歩行動作は重力に直行する方向への移動がメインであるのに対し,垂直移動動作は重力と平行な方向の移動である. 
	そのため動作中は壁から落ちないように壁を使って身体を支える必要がある. 

	この身体を支える方法に注目すると,壁を登るロボットは壁を掴むロボットと,壁に突っ張るロボットの２つに分類することができる.
	壁を掴んで移動するロボットとしてはLEMUR 3\cite{parness2017lemur3}やSCALER\cite{tanaka2022scaler}, LORIS\cite{nadan2024loris}がある. 
	これらは壁の突起を爪のついた脚先のグリッパでつかむことで, 壁面上でも身体を支えることができる. 
	一方でこれらは把持可能な突起がない平坦な壁を登ることができない. 
	また把持により身体を支える動作は平地での歩行では使わないため,歩行時には脚先にグリッパが邪魔になってしまうという欠点がある.

	\begin{figure}[t]
    \centering
      \includegraphics[width=0.95\columnwidth]{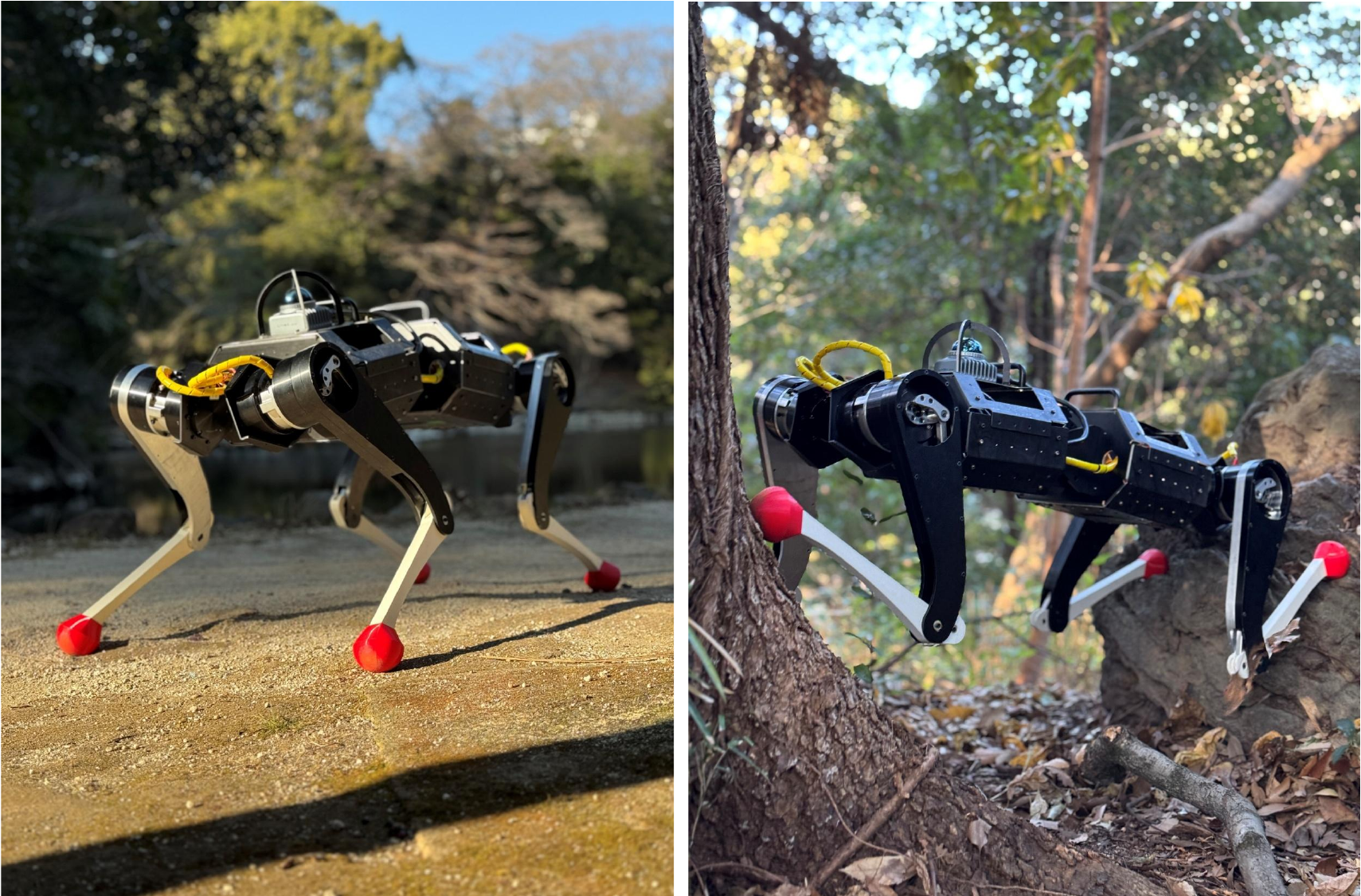}
      \vspace{-3.0ex}
      \caption{KLEIYNの概観. 歩行に加え,突っ張りを用いて壁登りを行う.}
      \vspace{-3.0ex}
      \label{fig:kleiyn}
    
  \end{figure}

	一方壁に突っ張る動作は, 歩行と比べると脚先にかかる力の向きが変わるだけであるため,歩行と同じ機構で実現でき, 壁に突起があるかどうかに関係なく,突っ張ることができる二枚の壁があれば壁面移動を行うことができる. 
	よって歩行と壁登りを両立するロボットには突っ張りを利用した壁登りが適していると考えられる.
	しかし四脚ロボットで壁に突っ張ることで垂直方向に移動を行うものは開発されておらず, 唯一存在する脚ロボットとしては六脚ロボットのSiLVIA\cite{lin2018twowall, lin2019twowall}がある.
	SiLVIAは二枚の壁に脚を突っ張ることで身体を固定し,脚を一本ずつ動かすことで壁面を移動することができる.

	しかしSiLVIAの動作には課題も多い.
	まずSiLVIAの壁登り動作は事前に計画した脚軌道にそって関節を動かす制御が主であり,脚が滑ってしまうとリカバリーができない.
	さらに関節を高減速比のモータで駆動しているため脚を動かす速度が遅く,壁面移動速度は3 mm/sと非常にゆっくりとしか移動できない.
	また胴体に関節がないため,胴体より狭い幅の壁は登ることができない. 
}%
{%
}%
\subsection{Locomotion on Legged Robots}\label{subsec:walking-robots}
\switchtext%
{%
	In recent years, RL-based control has significantly improved the locomotion performance of quadruped robots.  
	Hwangbo et al.\cite{hwangbo2019anymal} achieved unified locomotion control across various terrains by training a quadruped robot in simulation on flat ground, steps, and slopes.  
	Furthermore, by randomizing model properties and introducing external disturbances, they realized robust control that is resilient to modeling errors and disturbances.
	Additionally, terrain curriculum improved learning efficiency by gradually increasing the step height during training\cite{rudin2022leggedgym}.  

	In order to learn robust and adaptive control, it is essential to utilize simulation environment.
	\textit{Teacher-Student Learning}\cite{lee2020anymal, miki2022anymal} is a learning framework that leverages information available in simulation to improve real-world performance.
	This approach enables the robot to infer terrain shape and other unobservable factors indirectly from time-series data, allowing for more adaptive control.  
	Furthermore \textit{Asymmetric Actor-Critic}\cite{pinto2017asymmetric, radosavovic2024learning} is a method that provides the Critic with observations available only in simulation.
	By stabilizing value estimation, this method improves overall learning stability.  
	Compared to Teacher-Student Learning, Asymmetric Actor-Critic has the practical advantage of requiring only a single-stage training process, making implementation simpler.
}%
{%
	近年, シミュレータ上での強化学習を用いた手法により四脚ロボットにおける歩行性能が大きく向上している. 
	\cite{hwangbo2019anymal}ではシミュレータ内の平地や段差, 坂で歩行動作を学習することにより多様な地形に対する統一的な歩行制御を実現した.
	また摩擦係数などのモデルの特性をランダム化すること,そして外力などの外乱を加えることでモデル化誤差や外乱に強いロバストな制御を実現している.
	さらに\cite{rudin2022leggedgym}では段差を徐々に大きくするカリキュラム学習を導入することで,効率的な学習を実現した.

	一方現実でロボットが得られる情報はシミュレータ内で得られる情報よりも限られており,シミュレータ内でしか得られない情報を学習にどのように利用するかが重要となる.
	\cite{lee2020anymal, miki2022anymal}では, シミュレータで得られる情報を使う学習と現実で得られる情報のみを使う学習の二段階の学習を行うTeacher-Student学習を導入している.
	この学習法によりセンサからは直接得られない地形形状などの情報を時系列情報から間接的に推論する機構が実現され, より適応的な制御が可能となった.	

	また\cite{pinto2017asymmetric, radosavovic2024learning}ではCriticにシミュレータでのみ入手が可能な観測を与えるAsymmetric Actor-Criticを導入している. 
	Asymmetic Actor-Criticでは価値の推定が安定することで学習全体が安定する.
	この手法はTeacher-Student学習に比べ学習が1段階で済み,実装が簡単であるという実用上の利点もある.
}%
{%
}%

\section{Design of A Quadruped Robot KLEIYN with an Active Waist} \label{sec:design}

\subsection{Overview of Mechanism}\label{subsec:design-overview}
\switchtext%
{%
	\figref{fig:kleiyn-design-overview} shows the overall design of KLEIYN.  
	KLEIYN is a quadruped robot with a total of 13 degrees of freedom (DOF), consisting of 3-DOF per leg and 1-DOF in the torso.  
	The robot weighs 18 kg, with a body length of 760 mm and a standing height of 400 mm.  

	The torso consists of two identical components: the \textit{front-body-link} and the \textit{back-body-link}.  
	Each body link has a box-shaped structure made of four aluminum sheet-metal plates, housing internal components such as the onboard PC and battery.  
	These two body links are connected by the \textit{waist-joint}, which allows pitch-axis bending.  

	The leg design is based on MEVIUS\cite{kawaharazuka2024mevius}, an open-source metal quadruped robot.  
	Each leg consists of three links: the \textit{scapula-link}, \textit{thigh-link}, and \textit{calf-link}, which are connected by three joints: the \textit{collar-joint}, \textit{hip-joint}, and \textit{knee-joint}.  
	All leg joints are actuated by motors with a 1:10 reduction ratio and a maximum torque of 25 Nm.  


	A comparison of KLEIYN's physical parameters with existing quadruped robots is shown in \tabref{tb:kleiyn-physical-params}.

	\begin{table}[htbp]
    \centering
    \caption{Comparison between KLEIYN and Existing Quadruped Robots}
    \begin{tabular}{c|c|c|c|c} 
			Robot & Mass & leg length & DOF & Max Torque \\ \hline \hline
			ANYmal\cite{hutter2016anymal}  & 30 kg & 250 mm & 12 & 40 Nm \\ 
			Mini-Cheetah\cite{katz2019minicheetah}  & 9 kg & 200 mm & 12 & 17 Nm \\ 
			MEVIUS\cite{kawaharazuka2024mevius}  & 15.5 kg & 250 mm & 12 & 25 Nm \\ 
			KLEIYN (This Study)  & 18 kg & 250 mm & \textbf{13} & 25 Nm \\  
    \end{tabular}
    \label{tb:kleiyn-physical-params}
    \vspace{-3.0ex}
  \end{table}
}%
{%
  \figref{fig:kleiyn-design-overview}にKLEIYNの設計の全体像を示す.
	KLEIYNは各脚に3自由度,胴体に1自由度の計13自由度を持つ四脚ロボットである.
	質量は18 kgで胴体の全長は760 mm, 四脚で立っているときの高さは400 mmである.
	胴体はfront-body-linkとback-body-linkの２つで構成され,この２つは全く同じ部品である.
	胴体のリンク一つはアルミの板金4つを組み合わせた箱型の形状をしており,内部にPCやバッテリなどを格納する.
	２つの胴体リンクはwaist-jointでつながっており,pitch軸周りに曲げることができる.

	脚はオープンソースの金属製四脚ロボットMEVIUS \cite{kawaharazuka2024mevius}を参考に設計されている.
	脚はscapula-link, thigh-link, calf-linkの３つのリンクをcollar-joint, hip-joint, knee-jointの３つの関節でつなげることで構成されている.
	脚の関節はすべて減速比が1:10で最大トルクが25 Nmのモータで駆動されている.
	
	四脚ロボットの脚関節の構成はANYmal\cite{hutter2016anymal}のように膝関節が逆向きに曲がるものと,Spot\cite{bostondynamics2017spot}のように膝関節が同じ向きに曲がるものの２つに分類できるが,本研究では壁に突っ張るためANYmalのように膝関節が逆向きに曲がる関節配置とした. 
	既存の四脚ロボットと比較したKLEIYNの物理パラメータを\tabref{tb:kleiyn-physical-params}に示した.

	\begin{table}[htbp]
    \centering
    \caption{KLEIYNと既存の四脚ロボットの比較}
    \begin{tabular}{c|c|c|c|c} \toprule
			Robot & Mass & leg length & DOF & Max Torque \\ \hline \hline
			Mini-Cheetah\cite{katz2019minicheetah}  & 9 kg & 200 mm & 12 & 17 Nm \\ 
			ANYmal\cite{hutter2016anymal}  & 30 kg & 250 mm & 12 & 40 Nm \\ 
			MEVIUS\cite{kawaharazuka2024mevius}  & 15.5 kg & 250 mm & 12 & 25 Nm \\ 
			KLEIYN (This Study)  & 18 kg & 250 mm & \textbf{13} & 25 Nm \\  
    \end{tabular}
    \label{tb:kleiyn-physical-params}
    \vspace{-3.0ex}
  \end{table}

}%
{%
}%
\subsection{Design of Waist Joint}\label{subsec:design-waist}
\switchtext%
{%
	\begin{figure}[t]
    \centering
        \includegraphics[width=0.9\columnwidth]{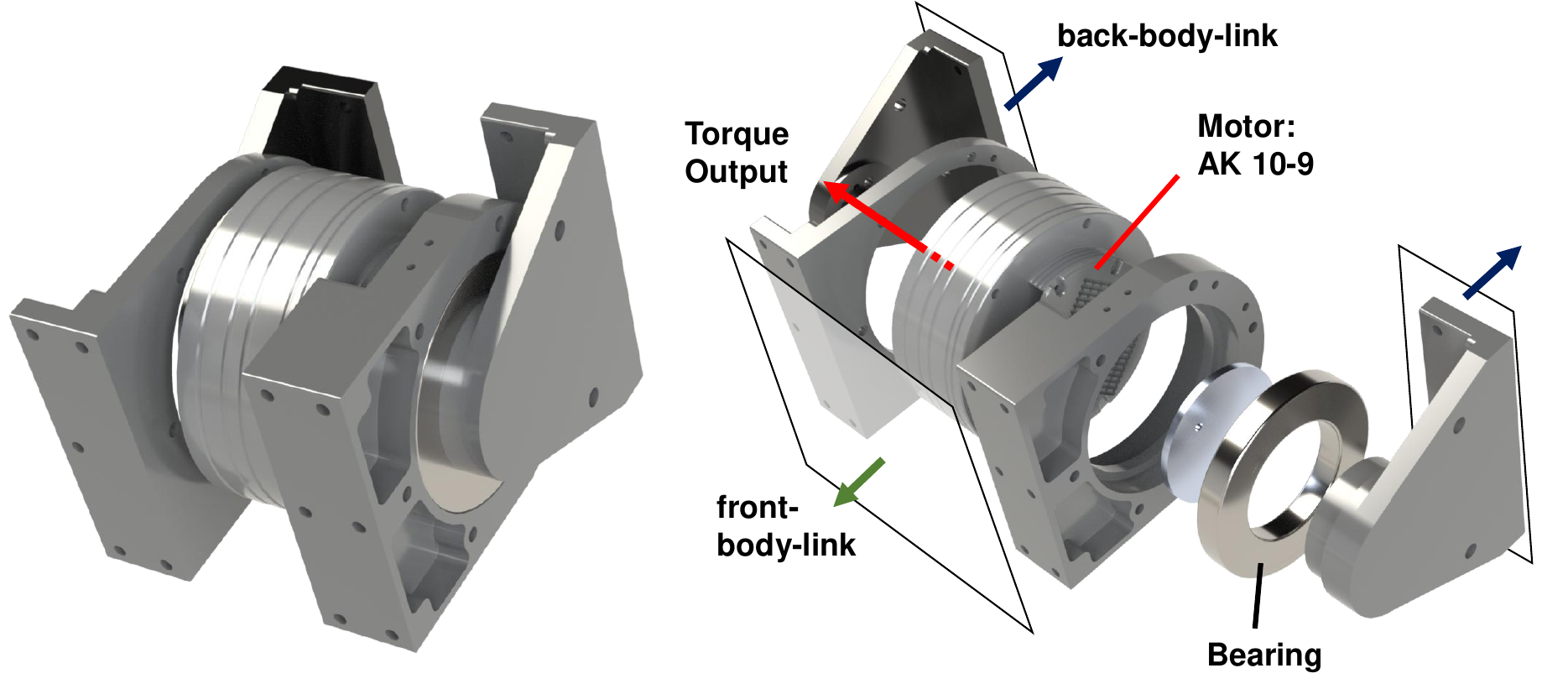}
        \vspace{-1.0ex}
        \caption{Design of KLEIYN's waist joint. The motor output is reduced by a 1:9 gear ratio inside the motor and transmitted to the rotation axis. To maintain rigidity, a bearing is placed on the opposite side of the output shaft to support the load.}
        \vspace{-3.0ex}
        \label{fig:kleiyn-waist-design}
    
	\end{figure}

	KLEIYN features a 1-DOF rotational joint along the pitch axis, allowing the torso to bend.  
	The design of the waist joint is shown in \figref{fig:kleiyn-waist-design}.  
	The frame is made of machined aluminum parts, and its strength was verified through pre-simulation stress analysis.  
	To enhance rigidity, a double-supported structure is adopted, where a bearing on the opposite side of the motor output shaft supports the load.  
	For easier sim-to-real transfer in reinforcement learning, the motor output is transmitted to the rotation axis via a low-reduction (1:9) gear in a quasi-direct-drive configuration.
}%
{%
	\begin{figure}[t]
    \centering
			\includegraphics[width=0.9\columnwidth]{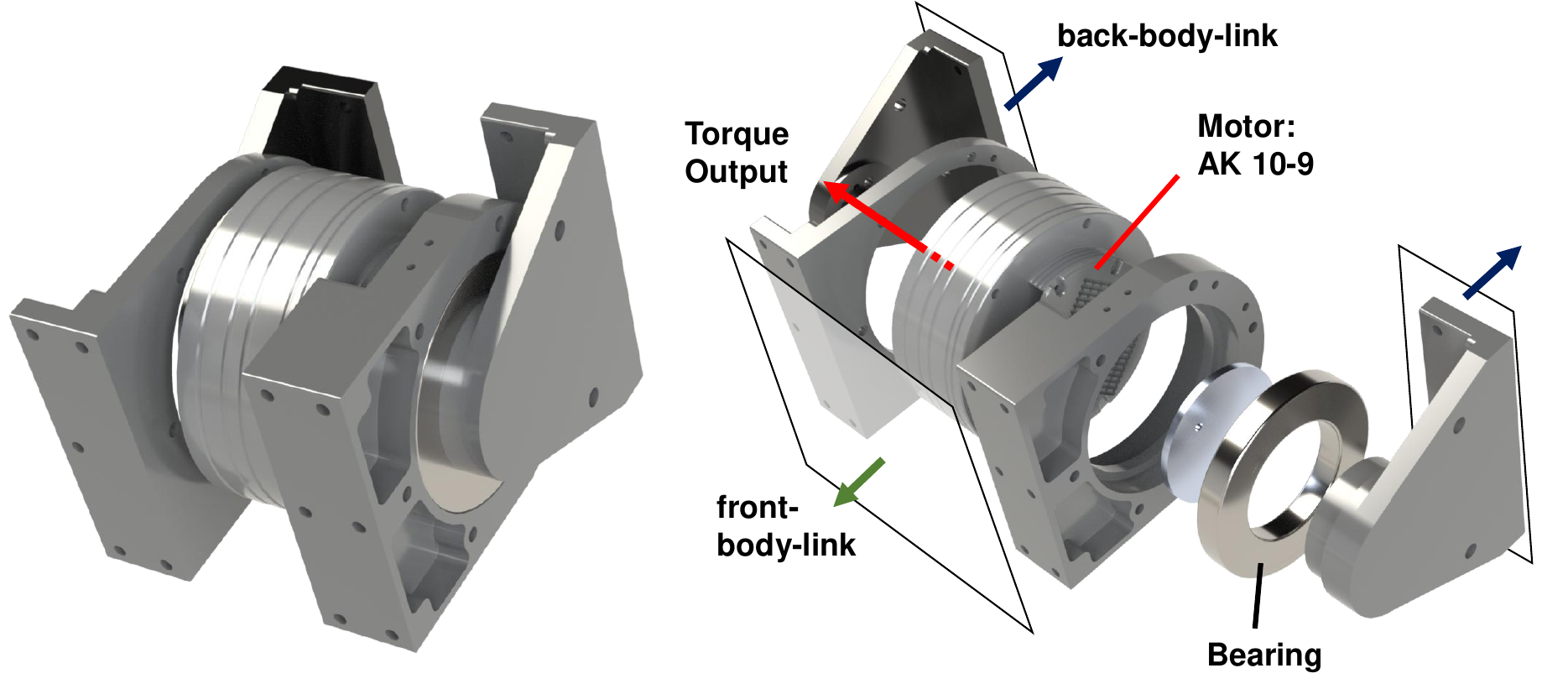}
      \vspace{-3.0ex}
      \caption{KLEIYNの腰関節の設計. 出力はモータ内部で1:10に減速され回転軸に伝わる. 剛性を保つために出力軸と反対側にベアリングを入れ荷重を受ける構造となっている.}
      \vspace{-3.0ex}
      \label{fig:kleiyn-waist-design}
    
  \end{figure}
	KLEIYNは胴体を曲げられるよう, pitch軸1自由度の回転関節を持つ.
	腰関節の設計を\figref{fig:kleiyn-waist-design}に示した.
	外骨格はアルミの切削部品で作られており事前に応力シミュレーションにより強度の確認を行っている. 
	またモータの出力軸の反対側はベアリンクで荷重を受ける両持ち構造とすることで剛性を高めている.
	強化学習におけるsim2realを容易にするため, モータの出力が1:9の低減速比ギヤを通じて回転軸に伝わるクアジダイレクトドライブ方式で駆動されている. 
}%
{%
}%

\subsection{Assessment of Motor Torque Suitability}\label{subsec:torque-assessment}
\switchtext%
{%
	\begin{figure}[t]
    \centering                                
        \includegraphics[width=0.95\columnwidth]{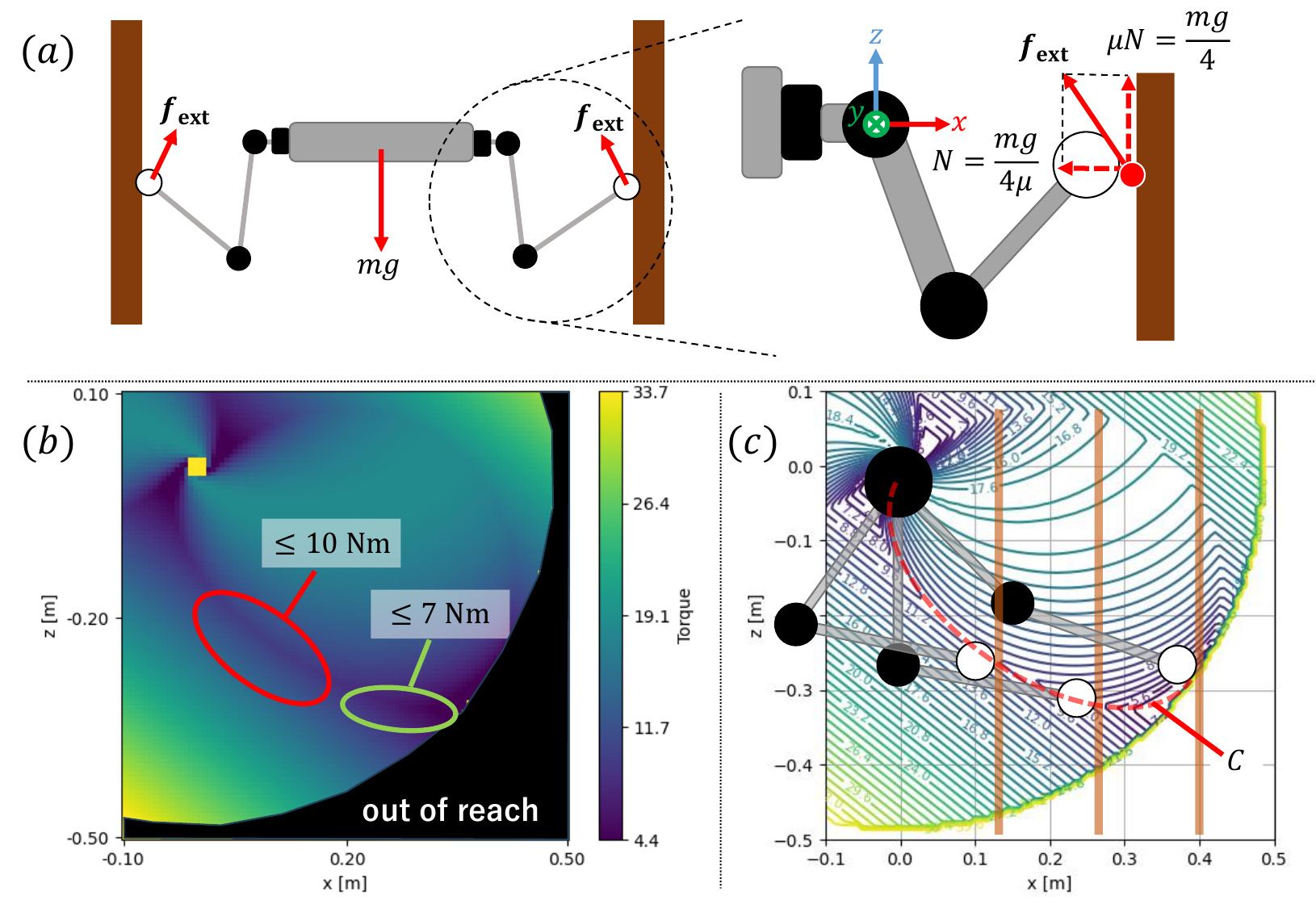}
        \vspace{-1.0ex}                             
        \caption{(a) The dynamic model of the chimney climbing used for torque estimation. The load during bracing is assumed to be evenly distributed across all legs, and the torque calculation is performed for a single leg. 
        (b) The maximum required torque for bracing at the foot position $\vectext[foot]{p}$ with external force $\vectext[ext]{f}$, calculated as $\tau_{max} = \max (\tau_{\{collar, hip, knee\}}) = \max (J(\vectext[foot]{p})^T \vectext[ext]{f})$.
        (c) Example foot positions using the curve $C$, where the torque is minimized.}                               
        \vspace{-3.0ex}                             
        \label{fig:kleiyn-torque-plot}
                                      
	\end{figure}                                    
        
	Bracing motion requires greater torque compared to walking.  
	To ensure stable operation, the required torque at each leg joint during bracing was calculated, and motor selection was based on this analysis.  

	The dynamic model used for the calculation is shown in \figref{fig:kleiyn-torque-plot}-(a).  
	Assuming a robot mass of 20 kg while neglecting the leg mass, the force applied to each foot during bracing is considered evenly distributed.  
	The coefficient of friction at the contact point was set to 0.8.  
	Each leg is modeled as a 3-DOF serial linkage, and the joint torques were computed using the Jacobian matrix $J(\vectext[foot]{p})$ at the foot position $\vectext[foot]{p}$ as:  
	\begin{equation}
			\vectext{\tau} = -J(\vectext[foot]{p})^T \vectext[ext]{f} \in \mathbb{R}^3.
	\end{equation}
	For simplicity, the collar-joint angle is assumed to be constant in calculation of $\vectext[foot]{p}$. The maximum torque values for different foot positions under these assumptions are plotted in \figref{fig:kleiyn-torque-plot}-(b).  

	From the figure, it is observed that despite variations in the distance between the wall and the hip joint, there exists a curve where the maximum torque locally decreases, denoted as curve $C$.  
	\figref{fig:kleiyn-torque-plot}-(c) illustrates curve $C$ along with examples of foot positions following this curve.  

	As shown in \figref{fig:kleiyn-torque-plot}-(b), if the foot position is adjusted along curve $C$, the maximum torque remains around 10 Nm, even with changes in the wall-to-hip distance.  
	Considering that the load may concentrate when other legs detach from the wall during actual operation, a maximum torque of approximately 20 Nm is deemed sufficient.  

	Since the waist joint experiences the combined torque from both left and right legs, it requires roughly twice the torque capacity of a single leg joint.  
	Thus, a motor with a maximum torque of about 40 Nm is adequate for the waist joint.  
	Based on these requirements, the AK 70-10 motor from T-MOTOR, with a maximum torque of 25 Nm, was selected for the leg joints.  
	For the waist joint, the AK 10-9 motor from T-MOTOR, which provides a maximum torque of 48 Nm, was chosen.
}%
{%
	\begin{figure}[t]
    \centering
			\includegraphics[width=0.95\columnwidth]{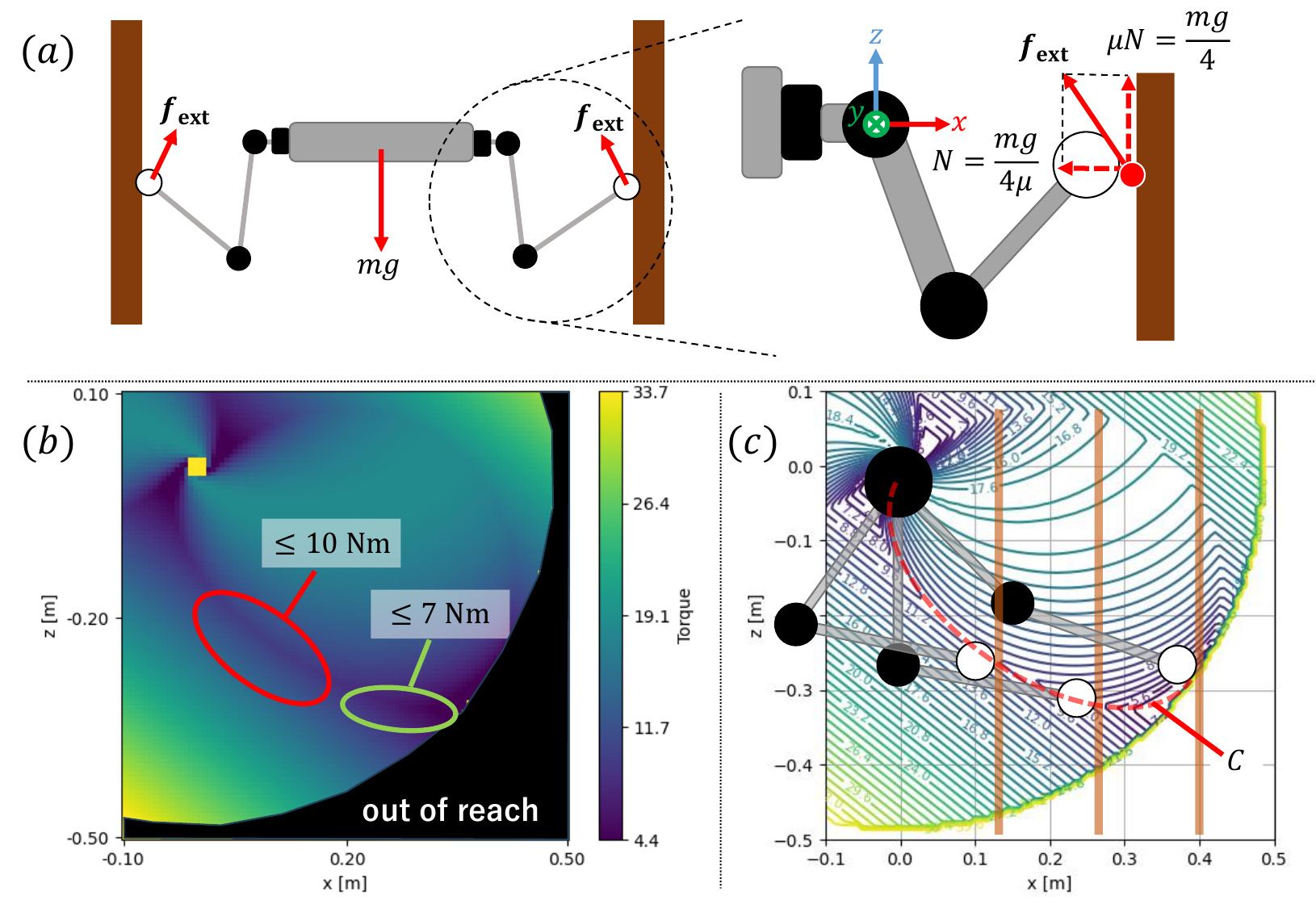}
      \vspace{-3.0ex}
			\caption{(a). トルクの見積もりに利用した突っ張り動作の力学モデル. 突っ張る際の荷重は各脚に均等にかかるとして一脚でのトルク計算を行った. (b). 脚先位置$\vectext[foot]{p}$において力$\vectext[ext]{f}$で突っ張る動作に必要な最大トルク$\tau_{max} = \max (\tau_{\{collar, hip, knee\}}) = \max (J(\vectext[foot]{p})^T \vectext[ext]{f})$をプロットしたもの. (c). トルクが小さくなる曲線$C$を利用した脚先位置の例}
      \vspace{-3.0ex}
      \label{fig:kleiyn-torque-plot}
    
  \end{figure}
	突っ張り動作は歩行動作に比べ必要なトルクが大きい動作である.
	そのため確実に動作を行えるように突っ張り時に脚関節で必要なトルクの計算とそれに基づくモータの選定を行った. 

	計算に使用した力学モデルを\figref{fig:kleiyn-torque-plot}-(a)に示した.
	ロボットの質量は20kgで脚の質量は無視した上で突っ張り時に各脚先に均等に力がかかるとし, 接触点の摩擦係数は 0.8 とした.
	脚の構造は三自由度のシリアルリンクであり脚先位置$\vectext[foot]{p}$に対応するヤコビ行列$J(\vectext[foot]{p})$を用いて各関節のトルク$\vectext{\tau} = -J(\vectext[foot]{p})^T \vectext[ext]{f} \in \mathbb{R}^3$を求めた.
	簡略化のためcollar-jointは動かないとし, この場合の脚先位置に対しトルクの最大値をプロットしたのが\figref{fig:kleiyn-torque-plot}-(b)である. 

	図から壁と脚の付け根の距離が変化しても,最大トルクが局所的に小さくなる曲線が存在することがわかり,これを曲線$C$とする. \figref{fig:kleiyn-torque-plot}-(c)に曲線$C$とそれを利用した脚先位置の例を示した. 

	\figref{fig:kleiyn-torque-plot}-(b)から壁と脚の付け根の距離が変化しても,脚先位置を曲線C上に調整すれば最大トルクは10 Nm程度に収まることがわかる.
	実際の動作時は他の脚が壁から離れることで負荷が集中することも考え,最大トルクが20 Nm程度あれば十分であると結論付けられる.

	腰関節は左右の脚のトルクを受けるため脚関節のモータの二倍程度のトルクが必要であり,最大トルク40 Nm程度のモータを選定すれば十分である.
	これらの条件をもとに脚関節には最大トルクが25 NmであるT-motor社のAK 70-10を, 腰関節には最大トルク 48 NmであるT-motor社のAK 10-9を選定した.
}%
{%
}%

\section{Reinforcement Learning-Based Control for Wall Climbing} \label{sec:rl}
\subsection{Control Framework}\label{subsec:control-framework}
\switchtext%
{%
	\begin{figure}[t]
    \centering                                
        \includegraphics[width=0.95\columnwidth]{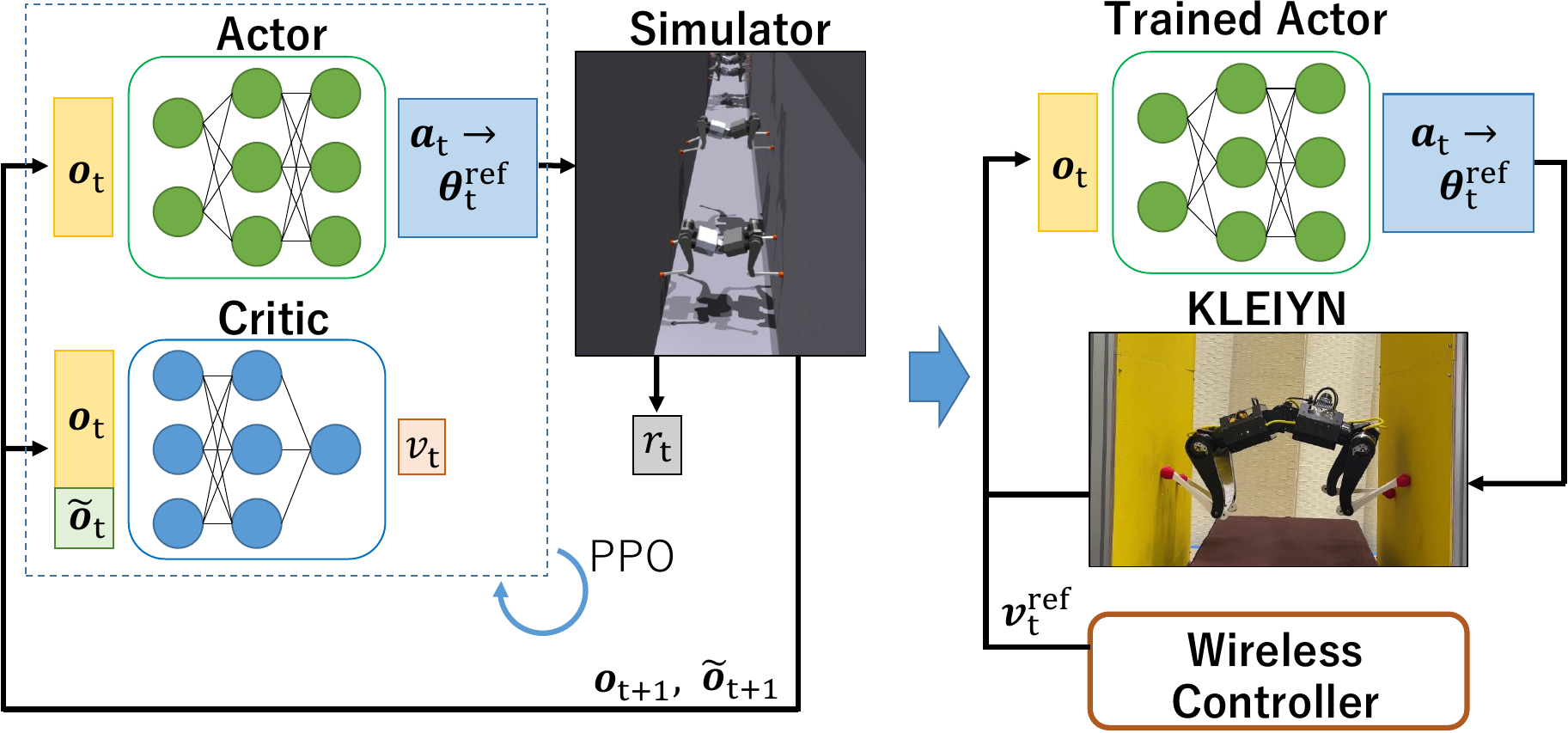}
        \vspace{-1.0ex}                             
        \caption{The training and real-world control framework of KLEIYN. During training, the Actor and Critic are learned using PPO \cite{schulman2017ppo} within the Asymmetric Actor-Critic framework. When applying the learned policy to the real robot, only the trained Actor is used, and control is performed to follow the commanded velocity input from the Wireless Controller.}
        \vspace{-3.0ex}                             
        \label{fig:kleiyn-control-framework}
                                      
	\end{figure}                                    
        
	The control framework of KLEIYN is shown in \figref{fig:kleiyn-control-framework}.  
	During training, two networks are trained: an Actor that outputs actions and a Critic that estimates value functions.  
	When operating the real robot, only the Actor network is used.  
	The Actor outputs scaled target joint angles at a frequency of 50 Hz.  
	These target joint angles are transmitted via CAN communication to each joint motor driver, where they serve as reference values for the internal position control of each motor driver.  
	KLEIYN's sensors include encoders in each motor, a 3D LiDAR (Livox MID-360) mounted on the front-body-link, and an IMU inside the LiDAR.  
	From the encoders, the joint angles $\vectext{\theta} \in \mathbb{R}^{13}$ and joint angular velocities $\vectext{\dot{\theta}} \in \mathbb{R}^{13}$ are obtained.  
	From the IMU, angular velocity $\vectext{\omega} \in \mathbb{R}^3$ and estimated orientation $\vectext{q} \in \mathbb{R}^4$ are acquired.  
	Additionally, a target velocity $v^{\mathrm{ref}}_{\mathrm{z}} \in \mathbb{R}$ is received from a Bluetooth controller.  
	The point cloud data obtained from the 3D LiDAR is used for SLAM with Fast-LIO \cite{xu2021fast}.  
	However, the localization results are only recorded and are not used for control.  
}%
{%
	\begin{figure}[t]
    \centering
      \includegraphics[width=0.95\columnwidth]{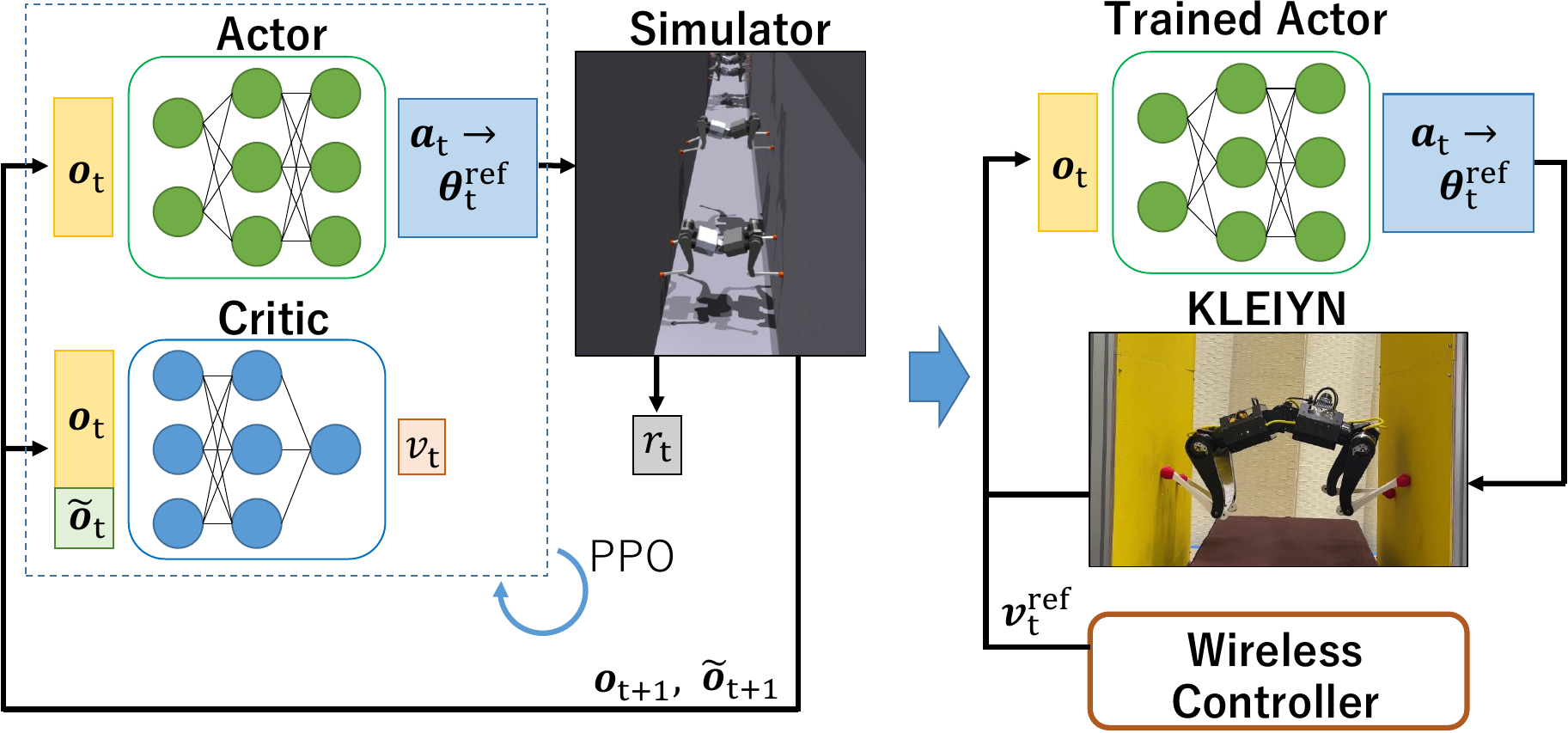}
			\caption{KLEIYNの制御フレームワーク. 状態$\mathbf{s_t}$から行動$\mathbf{a_t}$を出力するNNをシミュレータで学習する. 出力された行動は関節角度PD制御の目標値として使われる.}
      \vspace{-3.0ex}
      \label{fig:kleiyn-control-framework}
    
  \end{figure}
	KLEIYNの制御フレームワークを\figref{fig:kleiyn-control-framework}に示す.
	学習時には, 行動を出力するActorと推定価値を出力とするCriticの２つのネットワークを訓練し, 実機を動かす際にはActorのみを用いる. 
	Actorはスケーリングされた目標関節角度を50Hzで出力する.
	その後CAN通信で各関節のモータドライバに送られた目標関節角度は各モータドライバ内部の位置制御の目標値に使われる.
	KLEIYNのセンサとしては各モータのエンコーダと,front-body-linkに取り付けられた3DLiDAR, Livox MID-360内部のIMUを利用している.
	エンコーダからは関節角$\vectext{\theta} \in \mathbb{R}^{13}$と関節角速度$\vectext{\dot{\theta}} \in \mathbb{R}^{13}$が,IMUからは角速度$\vectext{\omega}\in\mathbb{R}^3$と推定姿勢$\vectext{q}\in\mathbb{R}^4$が得られる.
	さらにBluetoothで接続したwireless controllerから目標速度$\vectext{v}^{\mathrm{ref}}\in\mathbb{R}^2$が入力される.
	なお3D LiDARから得られた点群はFast-lio\cite{xu2021fast}による自己位置推定に用いているが,推定結果は動作の記録のみに用いており制御には一切利用していない.
}%
{%
	\begin{figure}[t]
    \centering
      \includegraphics[width=0.6\columnwidth]{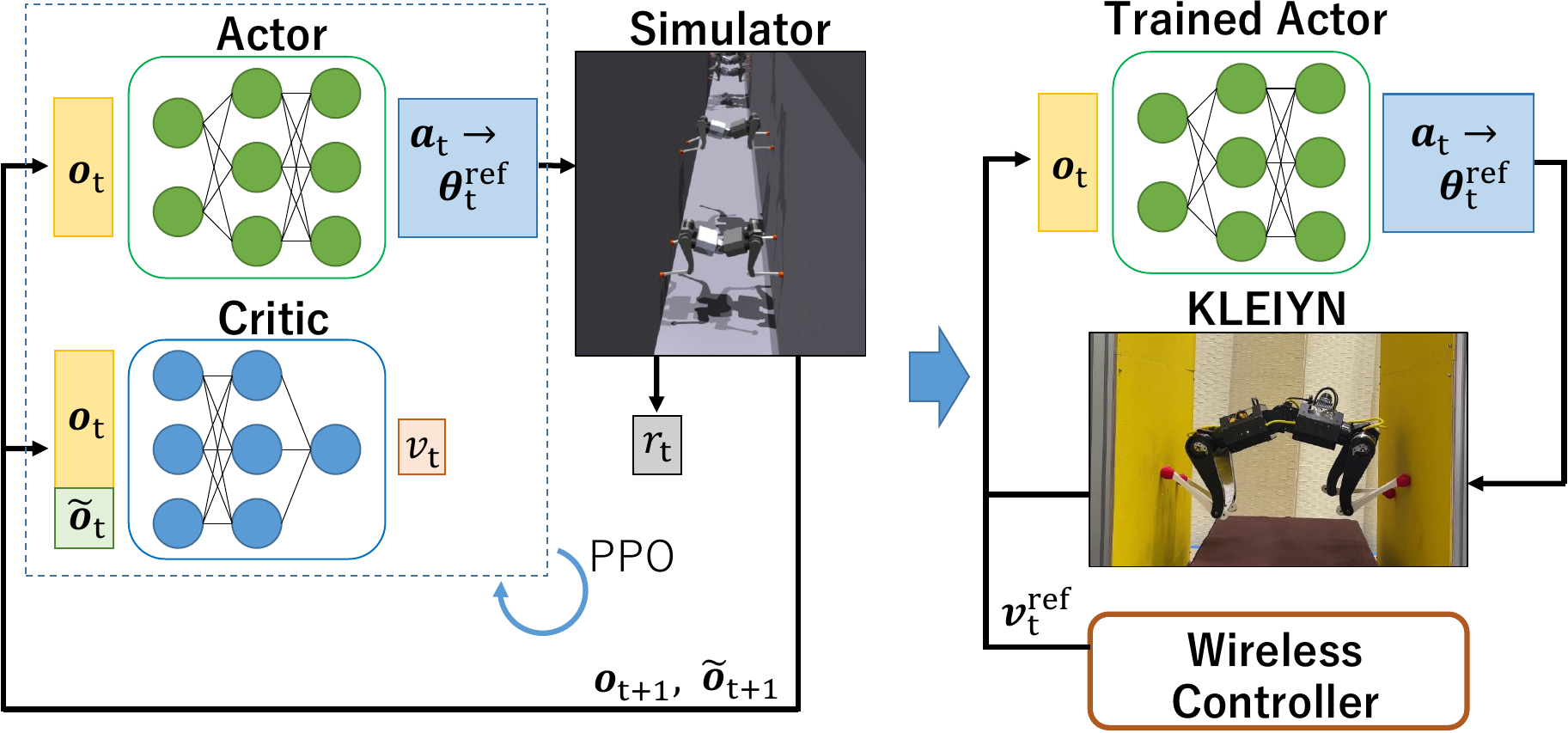}
			\caption{KLEIYNの制御フレームワーク. 状態$\mathbf{s_t}$から行動$\mathbf{a_t}$を出力するNNをシミュレータで学習する. 出力された行動は関節角度PD制御の目標値として使われる.}
      \vspace{-3.0ex}
      \label{fig:kleiyn-waist-design}
    
  \end{figure}
	* stateからactionを出力するActorと,state+privileged informationから推定価値を出力するCriticをシミュレータ内で学習する.
	* 実機ではActorのみを用い, actionはスケーリングにより目標関節角度に変換され,各関節ごとに角度に対するPD制御が行われる.
	* 学習の際にはタスクに合わせた環境と報酬を設定し,経験をサンプリングし学習が行われるモジュールは同じものを用いた.
	* シミュレータはIsaac Gym, 学習アルゴリズムはPPOを用いた.
}%

\subsection{Reinforcement Learning via CGCL for Efficient Learning}\label{subsec:wall-climbing}
\subsubsection{Contact-Guided Curriculum Learning (CGCL)}\label{subsubsec:contact-guided-curriculum-learning}  
\begin{figure}[t]  
			\centering  
			\includegraphics[width=0.95\columnwidth]{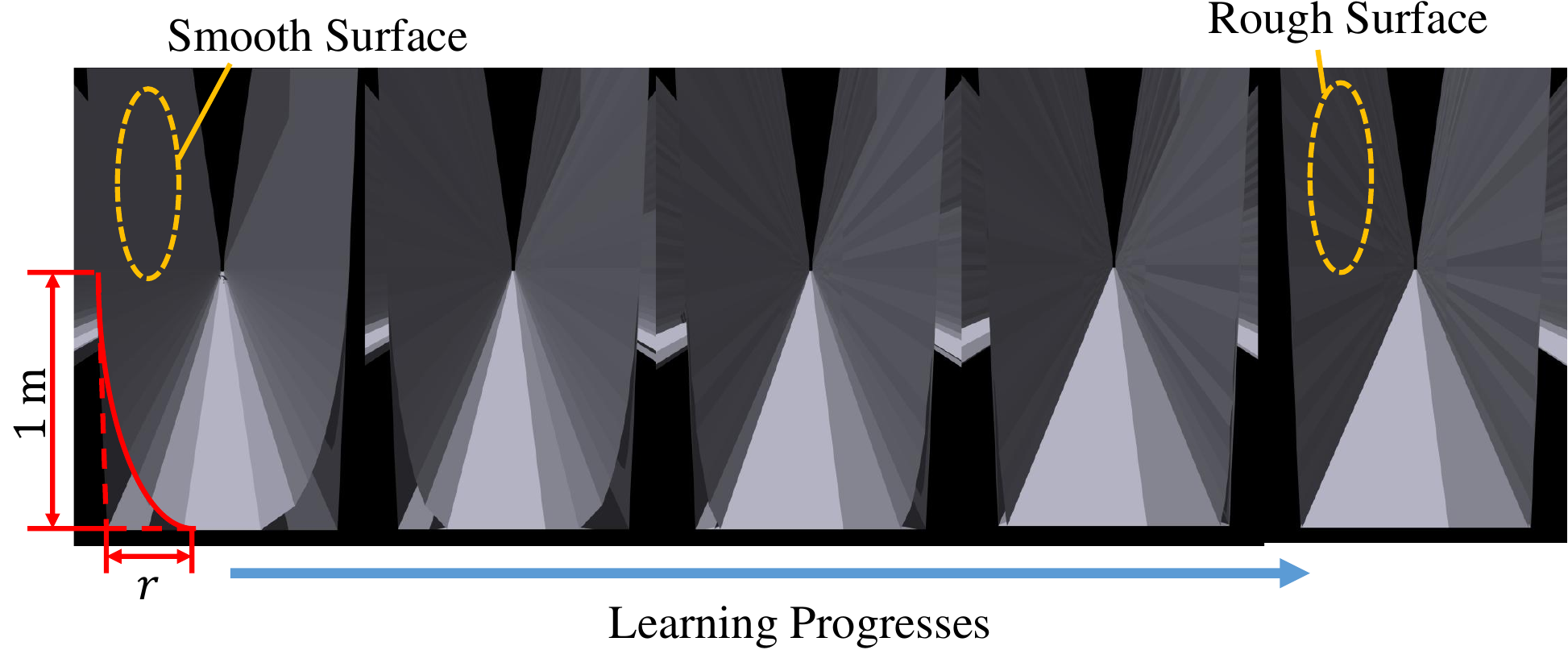}  
			\vspace{-1.0ex}  
			\caption{The terrain used for wall climbing learning. Initially, the floor has a U-shape, which gradually transitions to a vertical surface while the roughness of the wall increases.}  
			\vspace{-3.0ex}  
			\label{fig:kleiyn-climbing-terrain}  
	\end{figure}

	The key feature of chimney climbing learning is CGCL, where the junction between the wall and the floor gradually transitions from a curved surface to a vertical surface, as shown in \figref{fig:kleiyn-climbing-terrain}.  
	Since this study aims to achieve both horizontal and vertical movement, it is desirable for vertical movement to start from a standing position on the ground.  
	However, transitioning from a standing posture to a bracing posture requires the agent to experience receiving rewards for successful bracing, which can take a considerable amount of time due to the random exploration nature of the learning algorithm.  
	To accelerate learning, we designed an environment where, in the early stages of training, the terrain naturally induces the agent to experience bracing by randomly moving its feet.  

	For this purpose, we utilized a curved junction represented by an elliptical arc with a vertical radius of 1.0 m and a horizontal radius of $r$.  
	By adjusting $r$, the shape can be made smoother or more vertical.  
	Initially, we set $r = 0.3$ m and gradually reduced it to $r \rightarrow 0.0$ m as training progressed, enabling a smooth transition to fully vertical climbing.  
	Additionally, to ensure adaptability to non-flat wall surfaces, the roughness of the wall was increased as learning progressed.  

	\subsubsection{Learning Environment}\label{subsubsec:learning-environment}  
	Wall-climbing learning was conducted in a simulation environment using Isaac Gym \cite{liang2018isaacgym}.  
	To enable the robot to climb walls of various widths, we prepared walls ranging from 900 mm to 1100 mm in width and used them for training.  
	Additionally, to account for modeling errors, we randomized the coefficient of friction of the feet within the range of 0.7 to 0.95 and introduced variations in the mass and moments of inertia of each link.  
	To enhance robustness against external disturbances, we periodically applied random velocity perturbations of 0 to 1 m/s to the front-body-link and continuously applied random external forces and torques.  
	The initial position of the robot was set to a standing posture on the ground ($p_z = 0.4$ m), and the wall height in the simulator was set to 4 m.  
	If the robot successfully climbed the wall ($p_z \geq 3$ m) during training, it was rewarded and reset to the initial position, allowing for more efficient training cycles.  

	\subsubsection{Observation Definition}\label{subsubsec:state-definition}

	The learning algorithm employed Proximal Policy Optimization (PPO) \cite{schulman2017ppo}, and we introduced an Asymmetric Actor-Critic approach where different information was provided to the Actor and Critic.

	The observation for the Actor, denoted as \( \vectext[t]{s}=(\vectext[t]{o}) \), and the observation for the Critic, denoted as \( \vectext[t]{\tilde{s}} = (\vectext[t]{o}, \vectext[t]{\tilde{o}}) \), are defined as follows:

	\[
	\begin{aligned}
		\vectext[t]{o} &= (\vectext[t]{\omega}, \vectext[t]{g}, \vectext[t]{\theta}, \vectext[t]{\dot{\theta}}, v_{\mathrm{z}}^{\mathrm{ref}}) \\
		\vectext[t]{\tilde{o}} &= (d_t, \vectext[t]{n}, \vectext[t]{\tau}, \vectext[t]{p}, \vectext[t]{q}, \vectext[t]{\dot{p}}, \vectext[t]{\omega}, \vectext{\rho}, \vectext{m}, \vectext[ext]{f}, \vectext[ext]{\tau})
	\end{aligned}
	\]

	Here, $\vectext[t]{\omega}\in\mathbb{R}^3$ and $\vectext[t]{g}\in\mathbb{R}^3$ represent the angular velocity vector and the gravity direction vector in the robot's coordinate frame, respectively.  
	$\vectext[t]{\theta}\in\mathbb{R}^{13}$ and $\vectext[t]{\dot{\theta}}\in\mathbb{R}^{13}$ denote the joint angle vector and joint angular velocity vector, respectively, while $v_{\mathrm{z}}^{\mathrm{ref}} \in\mathbb{R}$ is the reference velocity.  

	Additionally, $d_t\in\mathbb{R}$ and $\vectext[t]{n}\in\mathbb{R}^3$ represent the distance to the nearest wall and its normal vector from the perspective of the front-body-link.  
	$\vectext[t]{\tau}\in\mathbb{R}^{13}$ represents the joint torque, and $\vectext[t]{p}\in\mathbb{R}$, $\vectext[t]{q}\in\mathbb{R}^4$, $\vectext[t]{\dot{p}}\in\mathbb{R}^3$, and $\vectext[t]{\omega}\in\mathbb{R}^3$ represent the global position, orientation, velocity, and angular velocity, respectively.  
	Furthermore, $\vectext{\rho}\in\mathbb{R}^{4}$ and $\vectext{m}\in\mathbb{R}^{18}$ represent the friction coefficients at the foot tips and the mass of each link, respectively.  
	$\vectext[ext]{f}\in\mathbb{R}^3$ and $\vectext[ext]{\tau}\in\mathbb{R}^3$ denote the external force and external torque applied to the front-body-link.  

	The origin of the robot, \( \vectext{p} = (p_{\mathrm{x}}, p_{\mathrm{y}}, p_{\mathrm{z}}) \), is placed at the hip joint, and the robot's orientation \( \vectext{q} \) is defined as the average orientation of the front-body-link and back-body-link.
	In learning, the target velocity \( v_{\mathrm{z}}^{\mathrm{ref}} \) was randomly assigned within the range of 0.0 m/s to 0.6 m/s.

	\subsubsection{Reward Definition}\label{subsubsec:reward-definition}

	The reward function was designed to encourage the robot's velocity \( \dot{p}_{\mathrm{z}} \) to follow the target velocity \( v_{\mathrm{z}}^{\mathrm{ref}} \) and to maintain a horizontal posture while moving along the wall.
	The rewards used in training are shown in \tabref{tb:climb-rewards}.

	\begin{table}[htbp]
	 \centering
	 \caption{Reward settings for wall climbing learning}
	 \begin{tabular}{c|c|c}\hline
			 Item & Weight & Definition  \\ \hline \hline
			 tracking velocity & 3.0 & $\mathrm{f}((\dot{p}_{\mathrm{z}}-v_{\mathrm{z}}^{\mathrm{ref}})^2+\dot{p}_{\mathrm{x}}^2, 0.01) $\\ 
			 collision & -1.0 & $\sum_{i\in P}(|\vectext[i]{f}^{\mathrm{contact}}| > 0.1)$\\ 
			 climb high & 5 & $p_{\mathrm{z}} > 3$\\
			 termination & -500 & $g_{\mathrm{z}} > 0.2$\\
			 orientation & -10.0 & $g_{\mathrm{x}}^2 + g_{\mathrm{y}}^2$ \\ 
			 rated torques & -3e-2 & $\vectext{\tau}/\vectext[rate]{\tau}$ \\ 
			 dof acc & -1e-6 & $||\vectext{\ddot{\theta}}||^2$ \\ 
			 dof vel & -3e-4 & $||\vectext{\dot{\theta}}||^2$ \\
			 dof pos limits & -10.0 & $||\vectext{\theta} - \mathrm{clip}(\vectext{\theta}, 0.8\vectext[min]{\theta}, 0.8\vectext[max]{\theta})||$ \\
			 dof vel limits & -0.1 & $||\vectext{\dot{\theta}} - \mathrm{clip}(\vectext{\dot{\theta}}, 0.6\vectext[min]{\dot{\theta}}, 0.6\vectext[max]{\dot{\theta}})||$\\
			 torque limits & -1e-3 & $||\vectext{\tau} - \mathrm{clip}(\vectext{\tau}, -0.8\vectext[max]{\tau}, 0.8\vectext[max]{\tau})||$\\
			 center & -0.1 & $p_{\mathrm{y}}^2$ \\
			 yaw & -1 & $\theta_{\mathrm{yaw}}^2$ \\
			 collar angles & -1 & $\sum_{i\in D}|\theta_{\mathrm{i}}^{\mathrm{collar}}|^2$ \\
			 low foot & 0.05 & $\sum_{i\in D}\log{(z_{\mathrm{i}}^{\mathrm{foot}}/0.2)}$\\
			 base height & 0.4 & $(p_{\mathrm{z}}-0.6) + 9.0 \ \min(p_{\mathrm{z}}-0.6, 0)$\\ \hline
	 \end{tabular}
	 \label{tb:climb-rewards}
	\end{table}

	Here, \( \mathrm{f}(x, \sigma) := \exp(-x^2/\sigma) - 0.6 x^2 \), and the index set \( P=\{ \)front-body-link, back-body-link, fl-scapula-link, fr-scapula-link, bl-scapula-link, br-scapula-link\( \} \) represents the links penalized for contact.
	The set \( D=\{\mathrm{fl, fr, bl, br}\} \) indicates the four legs.
	\( \vectext[i]{f}^{\mathrm{contact}}\in\mathbb{R}^{3} \) is the contact force vector applied to each link \( i \).
	\( (g_{\mathrm{x}}, g_{\mathrm{y}}, g_{\mathrm{z}}) \) are the components of the gravity direction vector \( {\mathbf{g}} \).
	\( \vectext[rate]{\tau} \in \mathbb{R}^{13} \) denotes the rated torques for each motor.
	\( \vectext[\{min, max\}]{\theta} \), \( \vectext[\{min, max\}]{\dot{\theta}} \), and \( \vectext[max]{\tau} \) represent the minimum and maximum values for joint angles, joint velocities, and joint torques, respectively.
	\( \theta_{\mathrm{yaw}} \) is the yaw angle of the robot's orientation expressed in Euler angles.
	Finally, \( \theta_{\mathrm{i}}^{\mathrm{collar}} \in \mathbb{R} \) and \( z_{\mathrm{i}}^{\mathrm{foot}} \in \mathbb{R}\) denote the collar joint angle and the z-coordinate of the foot position for each leg \( i \in D \), respectively.

\section{Experiments} \label{sec:experiments}

\subsection{Wall Climbing in Simulation}\label{subsec:exp-wall-climbing-sim}
\switchtext%
{%
	\subsubsection{Chimney Climbing for Wide Range of Walls}

	\begin{figure}[t]
			\centering
			\includegraphics[width=0.95\columnwidth]{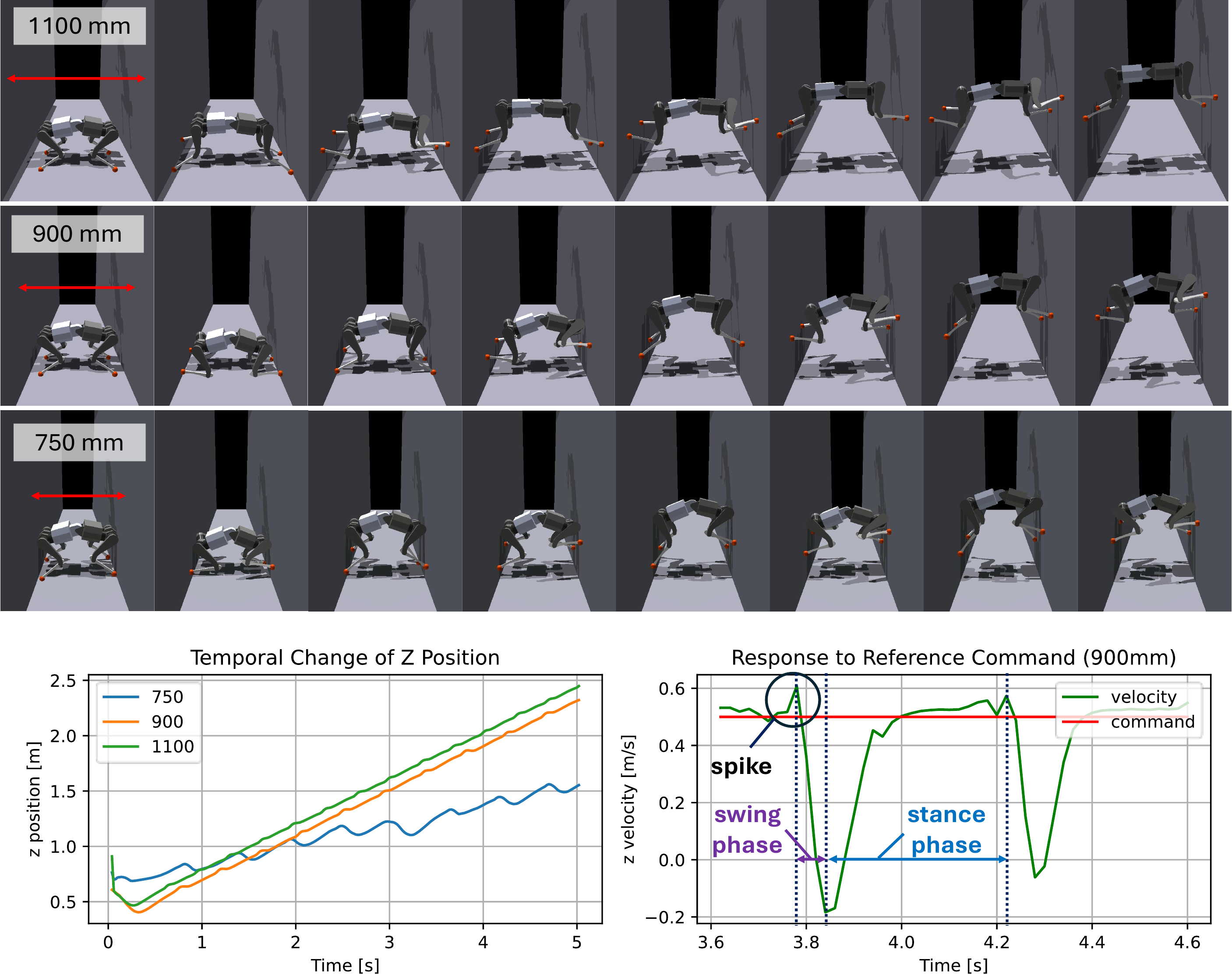}
			\vspace{-1.0ex}
			\caption{The wall-climbing experiment in the simulator. Top: climbing motion in the simulator for walls with a width of 750 mm, 900 mm, and 1100 mm at the target velocity $v_{\mathrm{z}}^{\mathrm{ref}} = 0.5 \mathrm{m/s}$. Bottom left: Time variation of the z-coordinate for different wall widths. Bottom right: Time variation of vertical velocity for the 900 mm-wide wall.}
			\vspace{-3.0ex}
			\label{fig:kleiyn-exp-climbing-sim-wide}
	\end{figure}

	Based on the learned policy, wall climbing motions were performed in the simulator for walls of width 750 mm, 900 mm, and 1100 mm at $v_{\mathrm{z}}^{\mathrm{ref}} = 0.5 \mathrm{m/s}$.
	\figref{fig:kleiyn-exp-climbing-sim-wide} shows the climbing motions, the variation of the z-coordinate over time, and the change in vertical velocity during the climbing motion on the 900 mm-wide wall.

	The climbing motion is realized by alternately executing two phases: the stance phase, where the body is pushed up, and the swing phase, where the legs are lifted.
	A significant upward velocity is momentarily generated in the vertical direction during the transition from the stance phase to the swing phase, indicating that the robot utilizes recoil when lifting its legs.
	Although the climbing speed decreases, the robot successfully climbs even the 750 mm-wide wall, which was not included in the training environment.
	This demonstrates the policy's generalization capability to unseen wall widths.
	Furthermore, as discussed in \secref{subsec:design-overview}, KLEIYN has a total length of 760 mm, making it impossible to climb a 750 mm-wide wall without utilizing its waist joint.
	This indicates that the robot has acquired the ability to climb narrower walls by leveraging its waist joint.

	\subsubsection{Effectiveness of CGCL}

	\begin{figure}[H]
			\centering
			\includegraphics[width=0.95\columnwidth]{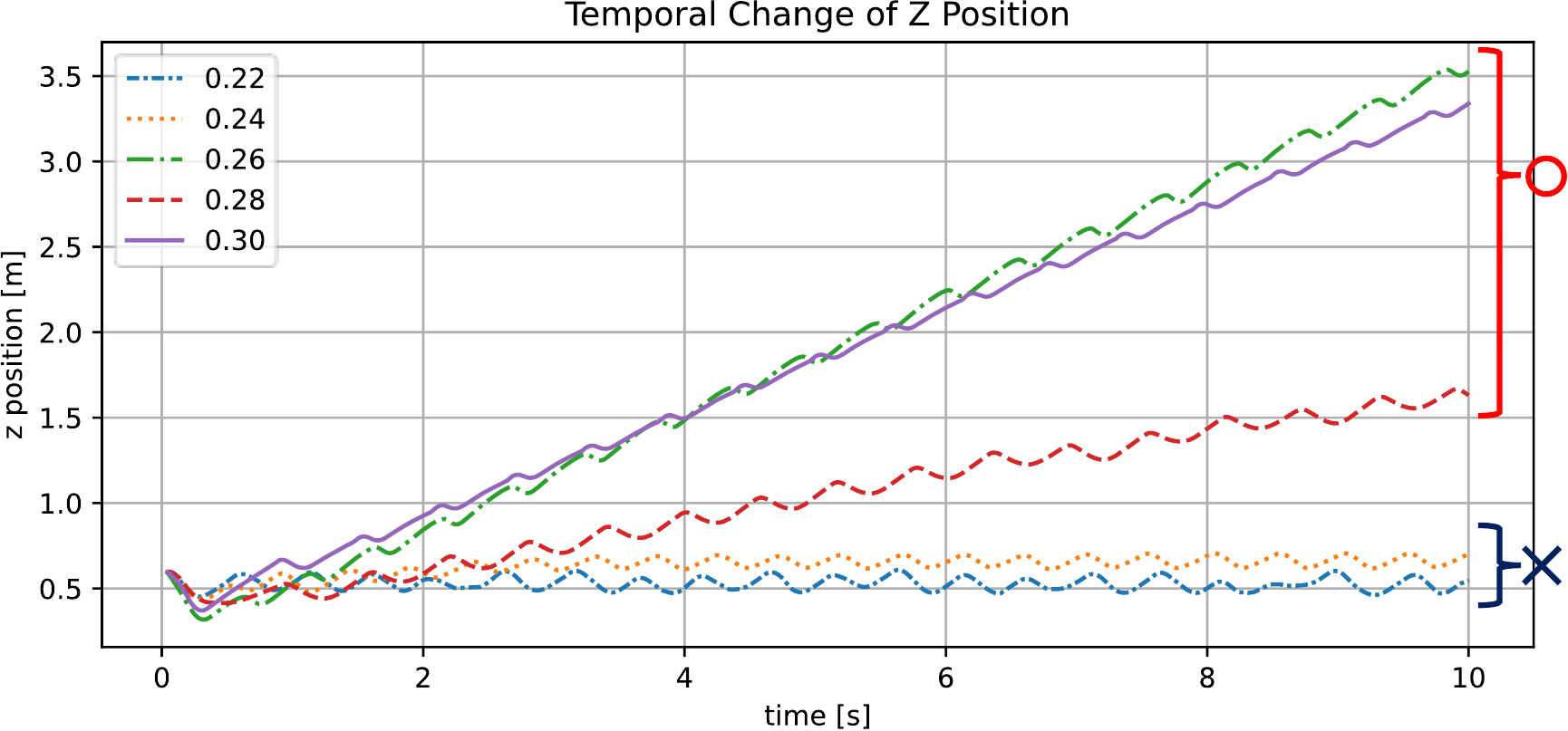}
			\vspace{-1.0ex}
			\caption{Comparison of models trained for 20,000 iterations with varying U-shaped parameter $r$. The graph shows time variation of the z-coordinate when commanding $v_{\mathrm{z}}^{\mathrm{ref}} = 0.5 \mathrm{m/s}$ on a 900 mm-wide wall.}
			\vspace{-1.0ex}
			\label{fig:kleiyn-exp-climbing-sim-r}
	\end{figure}

	To verify the effect of CGCL, the parameter $r$, as described in \secref{subsubsec:contact-guided-curriculum-learning}, was varied, and training was conducted for 20,000 iterations.
	For evaluation, a 900 mm-wide wall was prepared in the simulator, and an experiment was conducted where $v_{\mathrm{z}}^{\mathrm{ref}} = 0.5 $ m/s was commanded.
	\figref{fig:kleiyn-exp-climbing-sim-r} shows the variation of the z-coordinate over time.

	It was observed that the models successfully learned the climbing motion when $r=0.3, 0.28, 0.26$ m, whereas they failed to learn proper climbing behavior when $r=0.24, 0.22$ m.
	Furthermore, for values of $r$ smaller than 0.22 m, the models were completely unable to learn the climbing motion.
	These results suggest that CGCL effectively facilitates the learning of climbing motions and that a threshold exists around $r = 0.25$ m.

	\subsubsection{The Effect of Waist Joint on Tracking Performance}
	\begin{figure}[t]
			\centering
			\includegraphics[width=0.95\columnwidth]{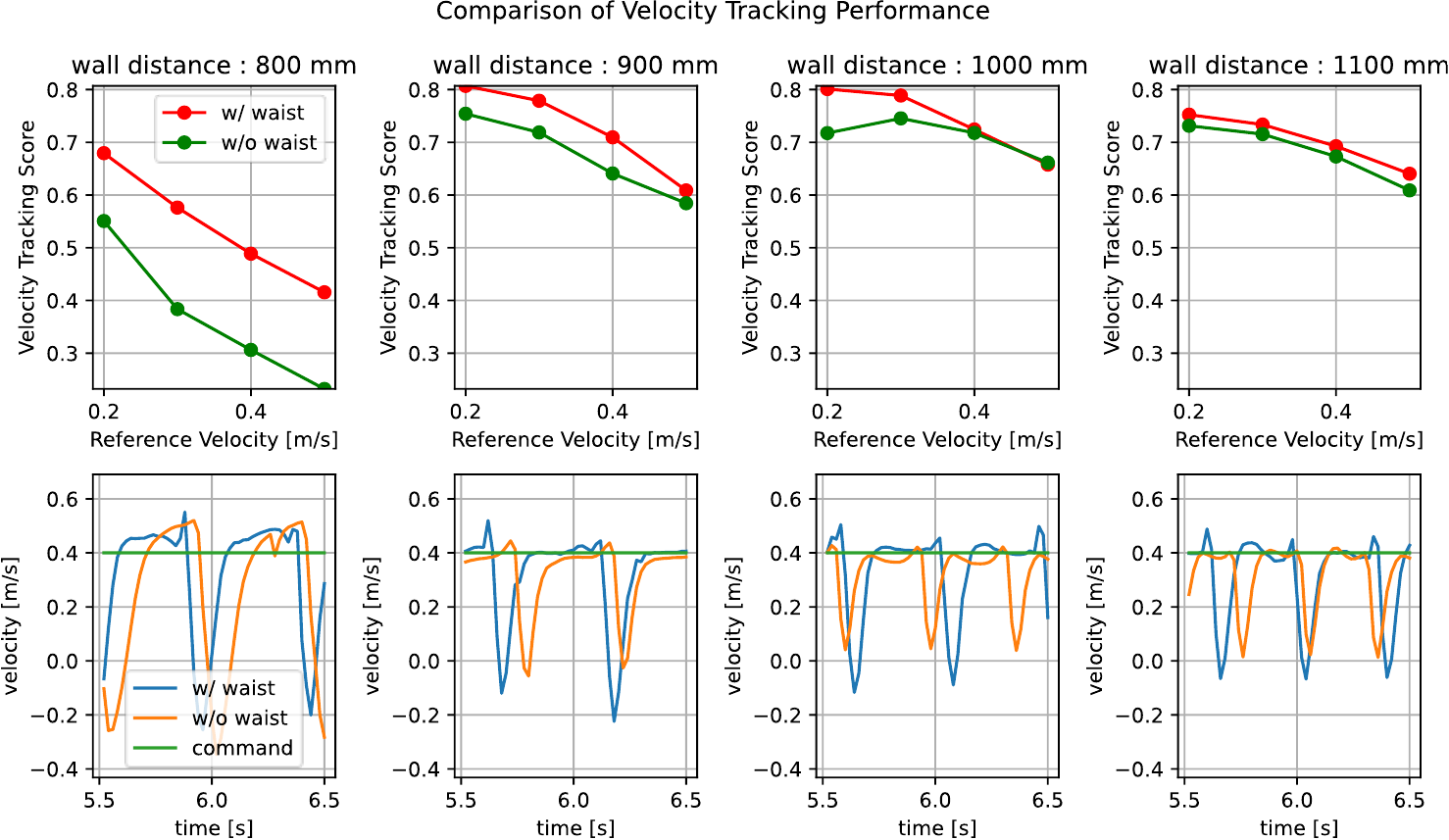}
			\vspace{-1.0ex}
			\caption{Comparison of tracking performance with and without the waist joint.  
			Top: Tracking score comparison when commanding a target velocity $v^{\mathrm{ref}}_{\mathrm{z}}$ in the range of $0.2 \sim 0.5$ m/s for 5 seconds on walls with a width of $800 \sim 1100$ mm.  
			Bottom: Comparison of velocity changes when commanding $v^{\mathrm{ref}}_{\mathrm{z}} = 0.4$ m/s on walls of width $800 \sim 1100$ mm.}
			\vspace{-3.0ex}
			\label{fig:kleiyn-vel-tracking-score}
	\end{figure}

	To investigate the impact of the waist joint on climbing, we conducted climbing experiments with models both with and without a waist joint.  
	For the model without a waist joint, training was conducted with the same reward and learning algorithm as the model with a waist joint, except that the waist joint was fixed.  
	\figref{fig:kleiyn-vel-tracking-score} shows the tracking score for the target velocity in both models.  
	The four graphs in the top section of the figure compare the tracking scores for walls with width of $800 \sim 1100$ mm when commanding $v^{\mathrm{ref}}_{\mathrm{z}} = 0.2 \sim 0.5$ m/s.  
	The bottom graph compares the velocity changes for walls with a width of $800 \sim 1100$ mm when commanding $v^{\mathrm{ref}}_{\mathrm{z}} = 0.4$ m/s, highlighting the motion differences between the models with and without a waist joint.  
	The tracking score was calculated as the mean value of $f(x, 0.01)$ used in the \textit{tracking velocity} reward function, as defined in \secref{subsubsec:reward-definition}.  

	\begin{figure*}[t]
    \centering
      \includegraphics[width=1.95\columnwidth]{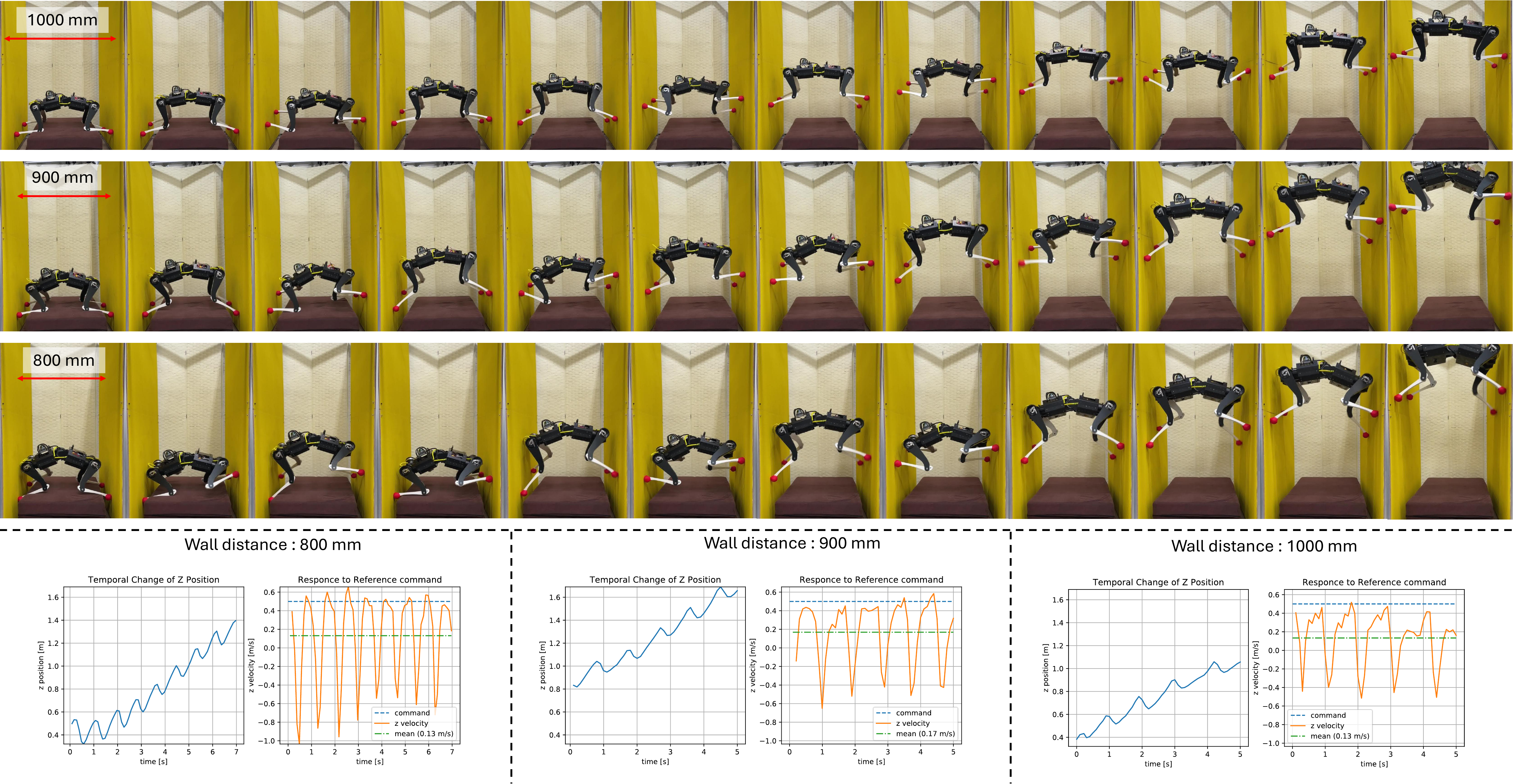}
      \vspace{-1.0ex}
      \caption{Wall-climbing experiments on walls with a width of 800 mm, 900 mm, and 1000 mm.  
      The figure shows the climbing motion and the corresponding time-series data of the z-coordinate and vertical velocity for each experiment.  
      In the 800 mm case, the robot achieved a maximum climbing height of 1.0 m with an average climbing speed of 170 mm/s.}
      \vspace{-3.0ex}
      \label{fig:kleiyn-exp-climbing}
	\end{figure*}
	
	The top graphs show that the model with a waist joint achieved higher tracking scores than the model without one.
	The difference in scores increased as the wall width decreased.  
	In the bottom graph, the model with a waist joint shows a sharper spike in velocity during the transition to the swing phase.
	This indicates greater utilization of recoil and it enhanced use of recoil likely contributes to improved tracking performance.  

	These results suggest that using the waist joint improves tracking performance, particularly on narrower walls, and facilitates motion that effectively utilizes recoil.  
}%
{%
	\subsubsection{Wall Climbing for Wide Range of Walls}

	\begin{figure}[t]
    \centering
      \includegraphics[width=0.95\columnwidth]{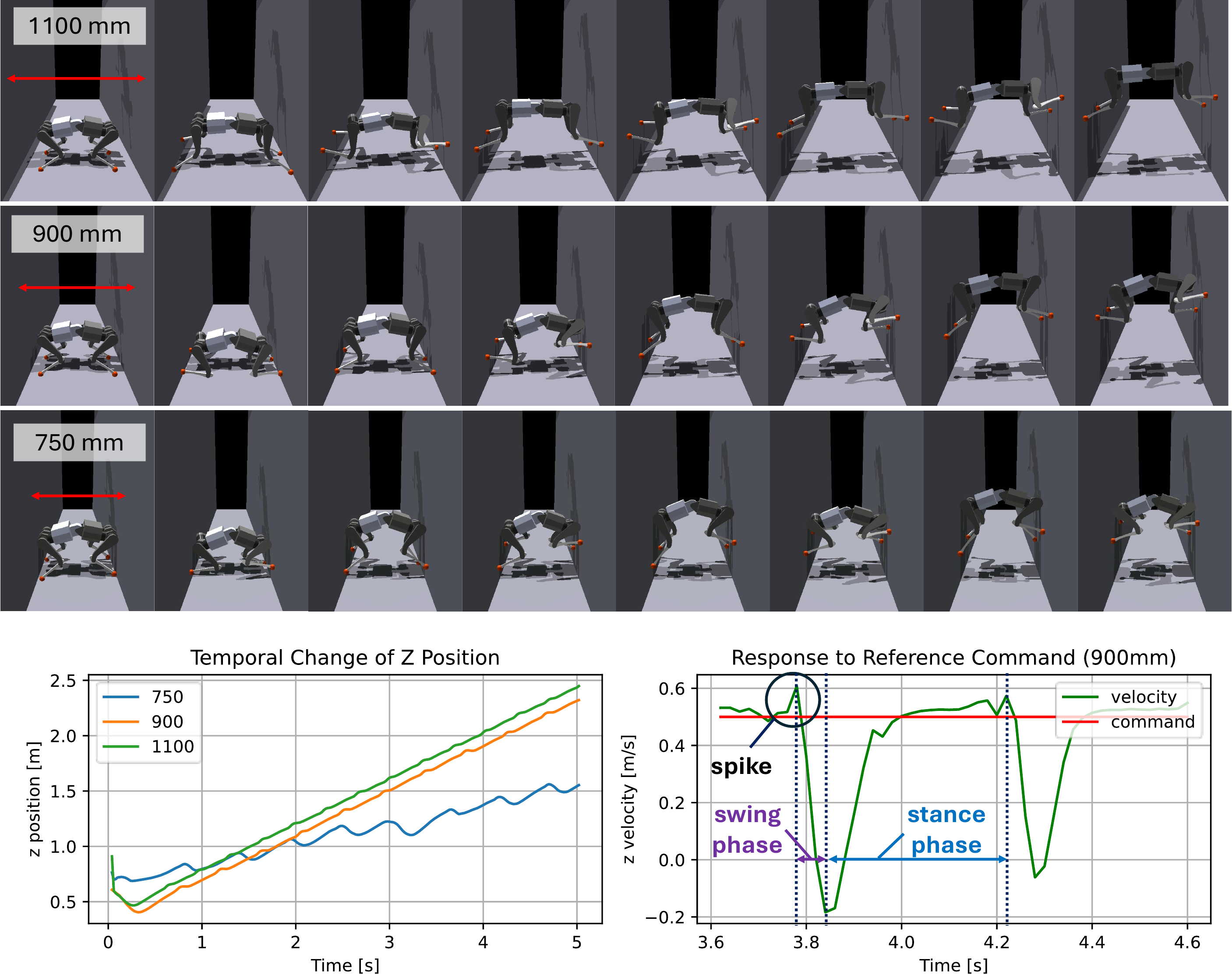}
			\caption{上：シミュレータにおいて幅750 mm, 900 mm, 1100 mmの壁に対し目標速度$v_{\mathrm{z}}^{\mathrm{ref}} = 0.5 \mathrm{m/s}$での壁登り動作の様子. 左下：幅 750 mm, 900 mm, 1100 mmでのz座標の時間変化. 右下: 幅 900 mm での鉛直方向の速度の時間変化.}
      \vspace{-3.0ex}
      \label{fig:kleiyn-exp-climbing-sim-wide}
    
  \end{figure}

	学習した方策に基づき, シミュレータ内で幅750 mm, 900 mm, 1100 mmの壁に対し,$v_{\mathrm{z}}^{\mathrm{ref}} = 0.5 \mathrm{m/s}$での壁登り動作を行った.
	壁登りを行った際の様子と,z座標及び 幅900 mm の壁における壁登り動作におけるz方向の速度の変化の様子を\figref{fig:kleiyn-exp-climbing-sim-wide}に示す.
	突っ張って体を持ち上げる動作 (stance phase) と脚を浮かせて上に引き上げる動作 (swing phase) の２つを交互に行うことで壁登り動作を実現している. 
	stance phaseからswing phaseに移行する際に一瞬鉛直方向に大きな上向きの速度が発生しており,脚を引き上げる際に反動を利用していることがわかる.

	また壁を登る速度は落ちるものの,学習環境には含まれていない幅 750 mm の壁でも壁登り動作を行うことができている.
	これは学習環境に含まれていない壁幅に対しても一定の汎化性能があることを示している.
	さらに \secref{subsec:design-overview}で述べたようにKLEIYNは全長760 mm であるため,腰関節を利用しないと幅750 mmの壁を登ることができない. 
	このことから腰関節を利用することでより幅の狭い壁を登る能力を獲得できていることがわかる.

	\subsubsection{Effectiveness of U-Shaped Wall}
	\begin{figure}[t]
    \centering
      \includegraphics[width=0.95\columnwidth]{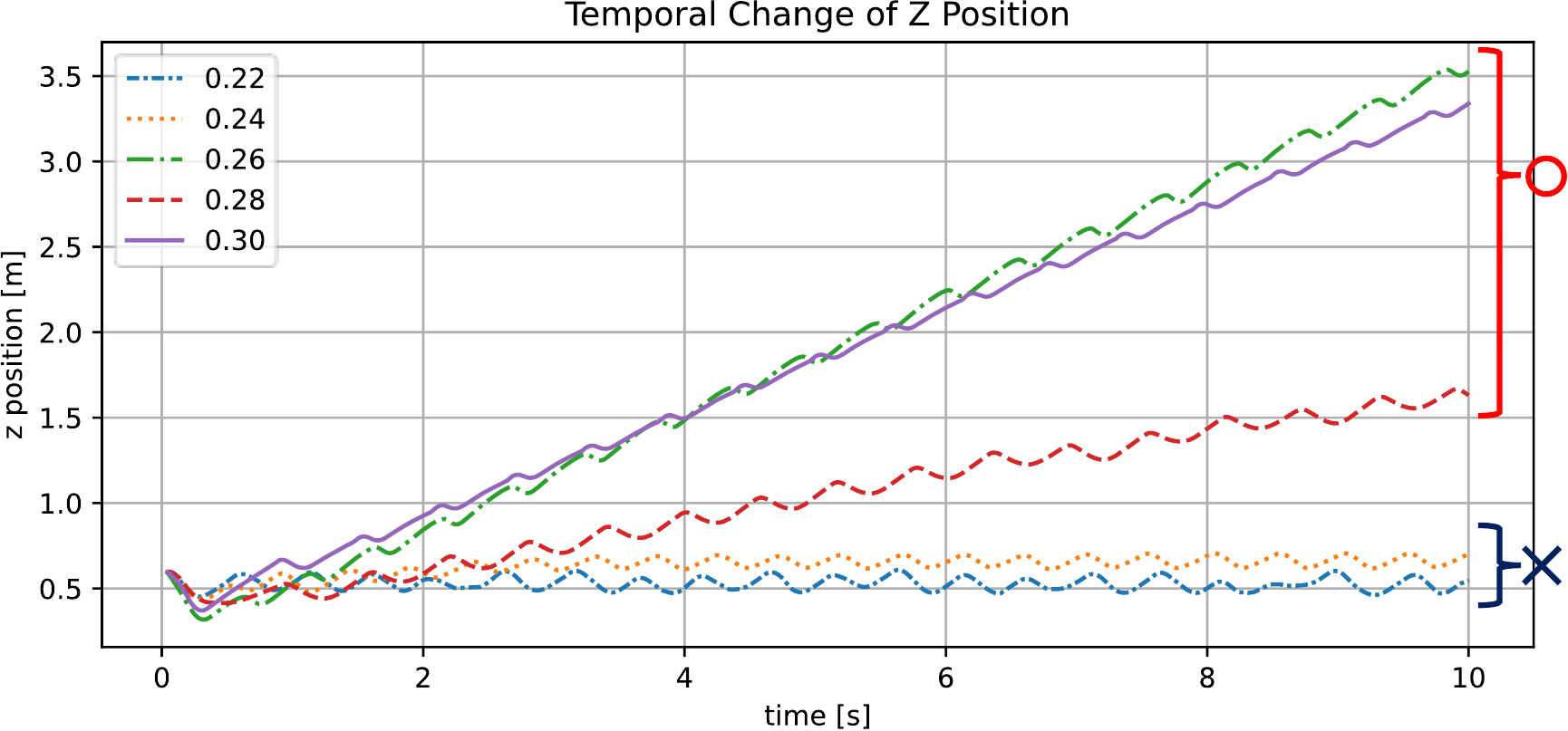}
			\caption{U字のパラメータ$r$を変化させて20,000 iteration学習を行ったモデルの比較. 左:壁の幅900 mmの壁において10 秒間$v_{\mathrm{z}}^{\mathrm{ref}} = 0.5 \mathrm{m/s}$を指令したときのz座標. 右:壁登りに成功したモデルにおける指令値と実際の速度の二乗誤差積分の値}
      \vspace{-3.0ex}
      \label{fig:kleiyn-exp-climbing-sim-r}
    
  \end{figure}

	壁と床の接合部を曲面にする効果を確かめるため, \secref{subsubsec:contact-guided-curriculum-learning}で述べた曲面のパラメータ$r$を0.3 mから0.22 mの範囲で変化させて20,000 iterationの学習を行った.
	評価のため学習後のモデルに対しシミュレータ上で幅900 mmの壁を用意し,10秒間$v_{\mathrm{z}}^{\mathrm{ref}} = 0.5 \mathrm{m/s}$を指令する実験を行った. 
	そのときのz座標の変化と指令値との誤差の二乗積分の値を\figref{fig:kleiyn-exp-climbing-sim-r}に示す.

	壁と床の接合部が$r=0.3, 0.28, 0.26$ mでは壁登り動作を学習できていることがわかり,$r=0.24, 0.22 \mathrm{m}$では壁を登る動作を正しく学習できていないことがわかる.
	また壁登りに成功したモデルでは $r = 0.28$ mのときが最も二乗誤差が大きく,$r = 0.3, 0.26$ mのときは同じ程度の誤差であることがわかる. 
	このことから接触誘導型カリキュラム学習は壁登り動作の促進に有効であるが,パラメータと学習促進の度合いは一定ではないことがわかる.

	\subsubsection{The Effect of Waist-Joint in Tracking Performance}

	\begin{figure}[t]
    \centering
      \includegraphics[width=0.95\columnwidth]{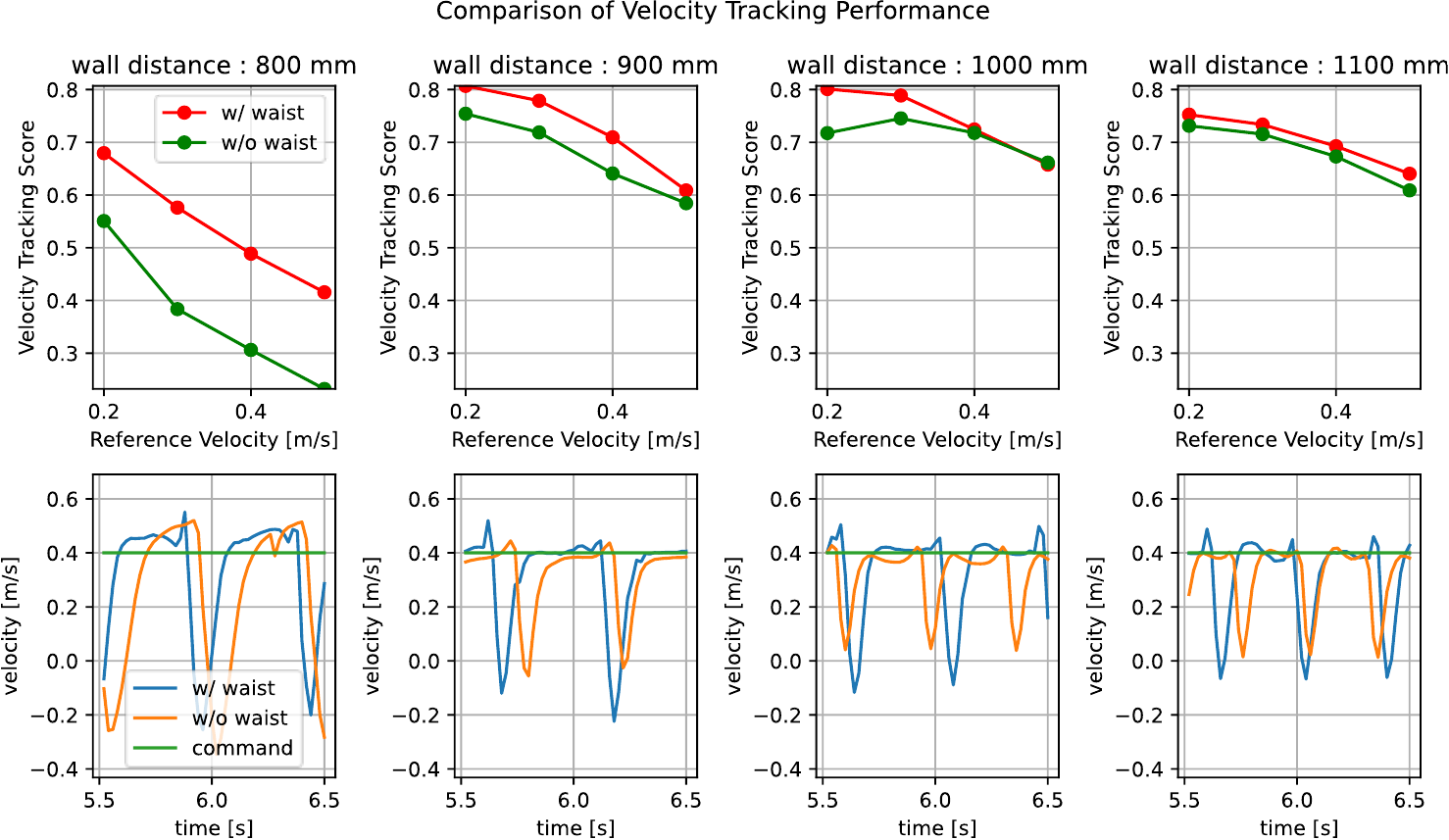}
			\caption{腰関節の有無による目標速度への追従性の比較. 上：幅$800 \sim 1100$ mm の壁に対し,目標速度$v^{\mathrm{ref}}_{\mathrm{z}}$を$0.2 \sim 0.5$ m/sの範囲で5 秒間与えた際の追従性のスコアの比較. 下：幅900 mm の壁に対し,目標速度$v^{\mathrm{ref}}_{\mathrm{z}} = 0.4$ m/sを与えた際の速度の変化を比較したもの}
      \vspace{-3.0ex}
			\label{fig:kleiyn-vel-tracking-score}
    
  \end{figure}
	
	腰関節の有無が壁登り動作に与える影響を調べるため,腰関節を持つモデルと持たないモデルで異なる幅の壁登り動作を行った.
	ここで腰関節を持たないモデルの学習は,腰関節を持つモデルの腰関節の角度を固定した上で報酬や学習アルゴリズムなどはすべて同じものを用いた.
	その際の目標速度への追従性を\figref{fig:kleiyn-vel-tracking-score}に示す.
	上の４つのグラフはそれぞれ幅$800 \sim 1100$ mm の四種類の壁に対し,目標速度$v^{\mathrm{ref}}_{\mathrm{z}}$を$0.2, 0.3, 0.4, 0.5$ m/sの四種類でそれぞれ5 秒間与えた際の追従性のスコアの比較を示している.
	また下のグラフは腰関節の有無による動きの違いを観察するため, 幅$800 \sim 1100$ mm の四種類の壁に対し,目標速度$v^{\mathrm{ref}}_{\mathrm{z}} = 0.4$ m/sを与えた際の速度の変化を比較したものである.
	ここで追従性を示すスコアは \secref{subsubsec:reward-definition}で定義した報酬関数の一部であるclimb\_tracking\_lin\_velで用いた $f(x, 0.01)$の平均値から算出している.

	上のグラフから腰関節を持つモデルは腰関節を持たないモデルに比べて目標速度への追従性のスコアが高く, 特に腰の有無によるスコアの差は壁の間隔が狭くなるほど大きくなっていることがわかる.
	また下のグラフの形状に注目すると腰関節を持つモデルはswing phaseに移行する際のspike状の速度変化が大きく,swing phaseにおける反動の動作がより大きいことがわかる.

	これらから腰関節を利用することで特に狭い壁における壁登り動作の追従性が向上し,反動を利用した動作を行いやすくなることがわかる.
}%
{%
}%

\subsection{Wall Climbing in the Real World}\label{subsec:exp-wall-climbing-real}
\switchtext%
{%
	To verify that the learned wall-climbing behavior could be applied to a real robot, we conducted physical experiments.  
	We set up a pair of plywood walls facing each other and tested climbing on walls of three different widths: 800 mm, 900 mm, and 1000 mm.  
	The robot was commanded to climb with a target velocity of $v^{\mathrm{ref}}_{\mathrm{z}} = 0.5$ m/s.  
	\figref{fig:kleiyn-exp-climbing} presents the experimental results, including the climbing motion and the time-series data of the z-coordinate and vertical velocity.  

	The results demonstrate that the robot successfully climbed in the commanded velocity direction for all wall widths, confirming that the learned model can generalize to different wall widths in the real world.  
	Notably, in the 800 mm case, the robot reached a maximum height of 1.0 m with an average climbing speed of 170 mm/s.  
	This corresponds to climbing 2.5 times its own height and achieving a climbing speed approximately 50 times faster than SiLVIA's reported average speed of 3 mm/s.  

	The climbing motion was achieved through alternating stance and swing phases, similar to the simulation experiments.  
	Furthermore, recoil-based movement, observed in the simulation, was also reproduced in the real-world experiments.  
	Additionally, slipping of the legs was frequently observed during the stance phase.  
	In such cases, the robot quickly repositioned the slipped leg and re-established bracing.  
	This suggests that the learned policy is robust against disturbances, such as slipping, through reinforcement learning.  

	However, differences between the simulation and real-world behavior were also observed.  
	First, the robot failed to climb a 1050 mm-wide wall due to insufficient torque.  
	Additionally, during the swing phase, the downward velocity was larger in real-world experiments than in the simulation.  
	These discrepancies are likely caused by differences in motor torque characteristics and frictional properties between the simulation and the real world.  
	Incorporating these factors into the learning process may further improve performance.  
	Another observed issue was that, despite being commanded only in the vertical direction, the robot moved horizontally and reached the edge of the wall, leading to falls.  
	This is because the current model lacks sensory perception of its environment and cannot detect when it reaches the wall's edge.  
	This issue could potentially be resolved by integrating environmental sensing, such as LiDAR, to provide the robot with spatial awareness.  
	Lastly, motor overheating posed a challenge for sustained real-world climbing.  
	During climbing, motor temperatures increased, occasionally triggering thermal protection and causing motor shutdowns.  
	The overheating consistently occurred in specific motors, suggesting that optimizing the learning process to distribute motor loads more evenly could help mitigate overheating and extend operation duration.  
}%
{%
	\begin{figure*}[t]
    \centering
      \includegraphics[width=2.0\columnwidth]{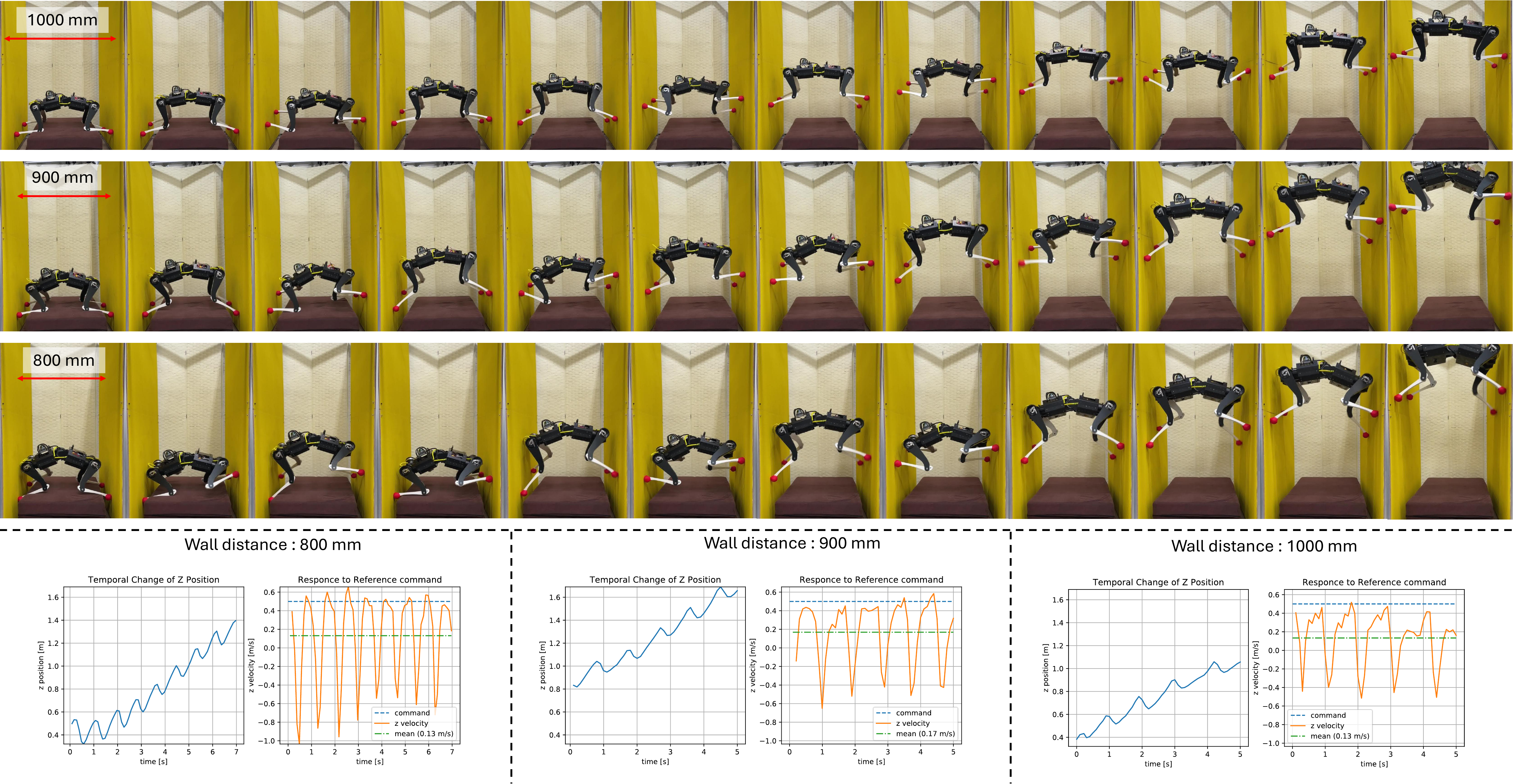}
			\caption{幅800 mm, 900mm, 1000 mm での壁面移動実験の様子と,それぞれの実験におけるz座標, z方向の速度の時間変化. 幅800 mm の場合は最大で高さ1.0 m, 平均速度170 mm/sの壁登りを達成している.}
      \vspace{-3.0ex}
      \label{fig:kleiyn-exp-climbing}
    
  \end{figure*}

	学習した壁面移動動作が実際のロボットに適用できることを示すため実験を行った.
	実験ではベニヤ板を向かい合わせに配置した壁を用意し,壁の幅を800 mm, 900 mm, 1000 mmの三種類に変化させ,目標速度$v^{\mathrm{ref}}_{\mathrm{z}} = 0.5$ m/sで壁登りを行った.
	実験の様子と,それぞれの実験におけるz座標, z方向の速度の時間変化を\figref{fig:kleiyn-exp-climbing}に示す.
	
	実験の結果,いずれの幅でも指令した速度方向に壁を登ることに成功し,学習したモデルが実機でも多様な幅の壁に対応できることが示された.
	特に幅800 mm の場合は最大で高さ1.0 m, 平均速度170 mm/sの壁登りを達成した. これは高さとしては自身の2.5倍の高さを登っており,さらにSiLVIAの壁登りの平均速度(3 mm/s) と比較し,約 50倍の速度で壁登りを実現している.

	また壁登りの動作はシミュレータで実験したときと同様にstance phaseとswing phaseの２つの動作を交互に行うことで実現しており,シミュレータで見られた反動を利用した動作などが実機でも再現されていることがわかる.

	一方でシミュレータと実機の動きには違いもあり, 1050 mmの壁に対してはトルク不足により壁を登ることができなかった.また実機ではシミュレータに比べswing phaseにおける下向きの速度が大きくなっていた.
	これらは実機のモータのトルク特性や摩擦特性のシミュレータとの違いによる出力不足が原因であると考えられ, 学習にこれらの特性を反映させることでさらなる性能向上が期待される.

	また今回は指令として垂直方向の目標速度を与えているだけであるが,実際には水平方向にも移動してしまい,壁の端まで移動して落ちてしまうことがあった. 
	これは現状のモデルは環境に対する感覚情報を持っておらず,壁の端に到達したことを検知できないためであるが,LiDARなどのセンサを用いて環境情報を取得することでこの問題を解決できると考えられる.

	さらに実機での壁登り動作の継続にはモータの発熱が問題となった. 壁登り中にモータの温度が上昇し,モータの保護機能によりモータが停止することがあった. 
	これらのモータの発熱は毎回特定のモータで発生しており,学習の際にモータの負荷を均等に分散させるような学習を行うことで発熱を抑えることができ持続時間を伸ばすことができると考えられる.
}%
{%
}%

\subsection{Locomotion in the Real World} \label{subsec:exp-locomoiton}
\switchtext%
{%
	\begin{figure}[t]
    \centering
      \includegraphics[width=0.95\columnwidth]{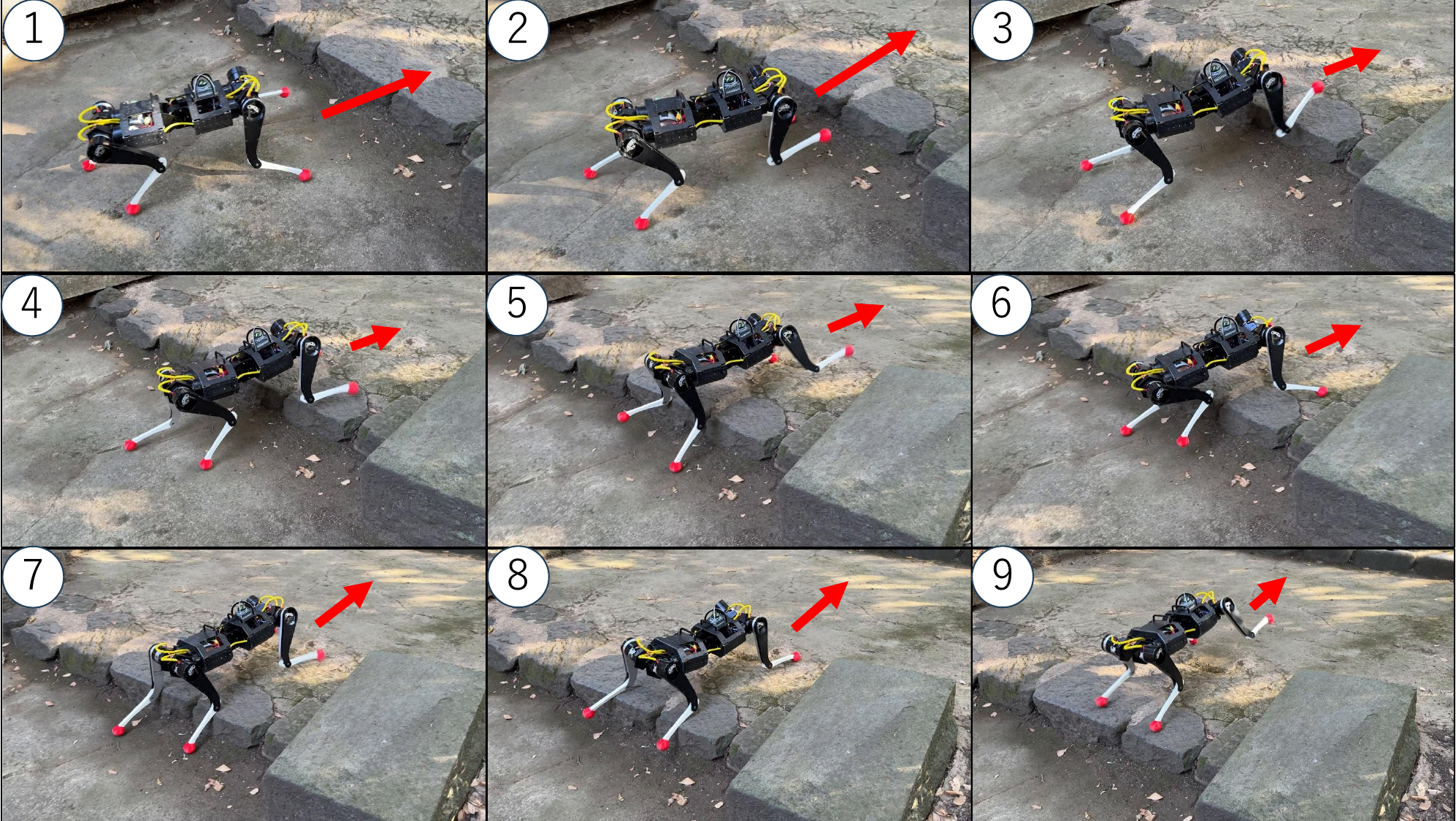}
      \vspace{-1.0ex}
      \caption{Outdoor locomotion experiments.  
      The robot successfully traversed a 150 mm step and maintained stability on slippery stones.}
      \vspace{-3.0ex}
      \label{fig:kleiyn-exp-locomotion}
    
	\end{figure}

	To demonstrate that KLEIYN is also capable of locomotion, we modified the wall-climbing reinforcement learning setup by using input commands  
	$\vectext{v}^{\mathrm{ref}} = (v_{\mathrm{x}}^{\mathrm{ref}}, v_{\mathrm{y}}^{\mathrm{ref}}, \omega_{\mathrm{z}}^{\mathrm{ref}})$ and providing rewards to encourage the robot to follow these reference velocities.  
	The trained locomotion policy was then deployed on the real robot for outdoor walking experiments.  
	\figref{fig:kleiyn-exp-locomotion} shows the robot's performance in these trials.  

	The test environment featured a staircase-like terrain with steps of approximately 150 mm in height.  
	KLEIYN successfully ascended and descended these steps using the learned policy.  
	Additionally, while walking on uneven stone surfaces, the robot occasionally experienced foot slippage.  
	In such cases, it was able to quickly recover by repositioning the slipping leg and restoring its posture.  

	These experiments confirm that the learned locomotion policy is effectively transferable to the real robot.  
	Furthermore, the results demonstrate that KLEIYN can successfully perform both wall climbing and walking, integrating these two modes of locomotion.  
}%
{%
	\begin{figure}[t]
    \centering
		\includegraphics[width=0.95\columnwidth]{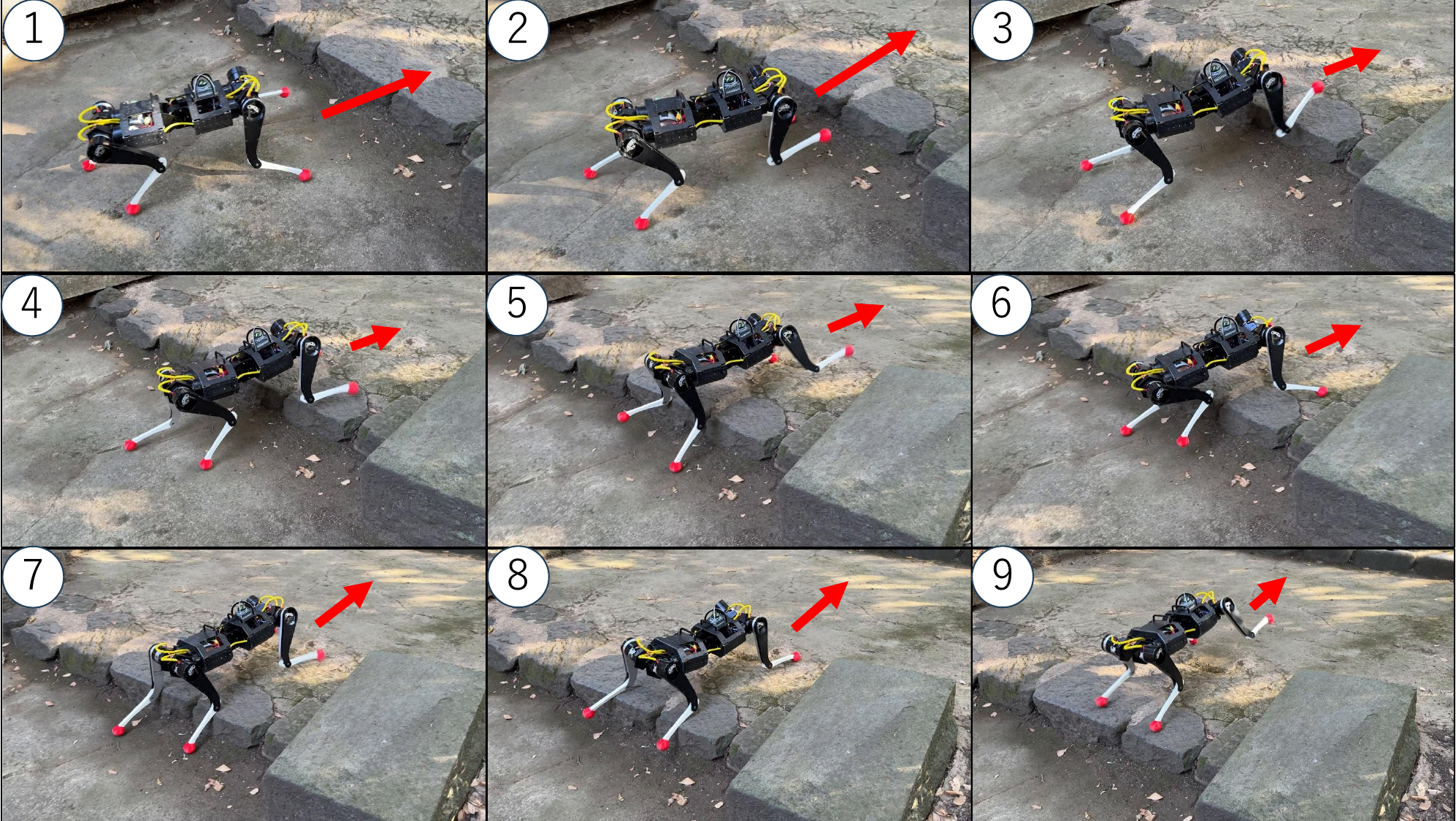}
		\caption{屋外環境での歩行実験の様子. 150mmの段差や滑りやすい石の上でも安定して歩行が行えている.}
		\vspace{-3.0ex}
		\label{fig:kleiyn-exp-locomotion}
  \end{figure}
	KLEIYNが歩行動作も可能であることを示すため, 壁面移動動作の学習において,入力を$\vectext[ref]{v} = (v_{\mathrm{x}^{\mathrm{ref}}}, v_{\mathrm{y}^{\mathrm{ref}}}, \omega_{\mathrm{z}^{\mathrm{ref}}})$とし,これらに追従するような報酬を与え, 学習を行った.
	学習した動作を実機に適用し屋外で歩行実験を行った際の様子を\figref{fig:kleiyn-exp-locomotion}に示した. 
	実験環境は高さ150 mm 程度の段差がある階段状の地形になっており,KLEIYNは学習した方策によってこれらの段差を上り下りすることに成功した. 
	また石の上を歩く際には脚が滑ることがあり,その場合も滑った脚をすばやく戻し,姿勢を立て直す動作を実現した.
	これらの実験を通して学習した歩行動作が実機にも正しく適用できていることが確認された.
}%
{%
}%

\section{Conclusion} \label{sec:conclusion}
\switchtext%
{%
	In this study, we developed KLEIYN, a quadruped robot with an active waist, capable of both walking and wall climbing.
	For wall climbing, we adopted the chimney climbing and introduced a waist joint to enable adaptation to wide range of walls.
	Through RL, KLEIYN successfully acquired wall-climbing skills and demonstrated its capability in real-world experiments across $800 \sim 1000$ mm of wall widths.
	To efficiently learn wall-climbing, we introduced a Contact-Guided Curriculum Learning, where the transition between the floor and the wall gradually changed from a smooth curve to a fully vertical surface.
	This method facilitated efficient learning by guiding the robot to bracing motion.
	Furthermore, our results demonstrated that the introduction of a waist joint significantly improved tracking performance, particularly on narrow walls.
	The waist joint allowed the robot to better utilize reactive motions, leading to enhanced adaptability and stability during climbing.
	This finding highlights the importance of waist joints in quadruped robot design and suggests potential applications beyond wall climbing, such as traversing complex three-dimensional environments.
}%
{%
	本研究では歩行と壁登りを同一のハードウェアで実現する四脚ロボットとしてKLEIYNを設計, 開発した.
	脚ロボットにおける壁登りの形態として突っ張り移動を採用し,多様な幅の壁に対応させるため腰関節を追加した.
	KLEIYNは強化学習により壁登り動作を学習し,シミュレータと実機双方で多様な幅の壁に対する壁登り動作を実現した.
	壁登り動作の学習において,従来歩行の学習で行われてきた地形をカリキュラム的に変化させる手法を派生させ, 壁と床の接合部をなめらかな曲面から徐々に垂直に変化させる地形支援的カリキュラム学習を行うことで壁登り動作を効率的に学習できることを示した.
	また腰関節を導入することで特に狭い幅の壁における壁登り動作の追従性が向上し,反動を利用した動作を行いやすくなることを示した. 
	この腰関節により動作の性能が向上するという事実は,四脚ロボットの設計において腰関節の重要性を示すものであり,壁登り以外の動作にも応用できる可能性がある.
}%
{%
}%

{
  \bibliographystyle{IEEEtran}
  \bibliography{bib}
}

\end{document}